\documentclass[journal]{IEEEtran}
\usepackage{amssymb}
\usepackage{graphicx}
\usepackage{textcomp}
\usepackage{amsmath}
\usepackage{color}
\usepackage{lineno}
\usepackage{url}
\usepackage{cite}
\usepackage{enumerate}
\usepackage{epstopdf}

\usepackage{graphicx}
\usepackage{graphics}
\usepackage{multirow}
\usepackage{caption}
\usepackage{subcaption}
\usepackage{gensymb}
\usepackage{float}

\usepackage{multirow}
\usepackage{array}
\usepackage{booktabs}
\usepackage{hhline}
\usepackage{xspace}
\usepackage{breqn}

\usepackage{algorithm}
\usepackage{algpseudocode}
\usepackage{soul}

\newcommand{\comst}[1]{}

\newcommand{\com}[1] {}

\newcommand{\ie}{\textit{i}.\textit{e}.,\@\xspace}
\newcommand{\eg}{\textit{e}.\textit{g}.\@\xspace}

\DeclareMathOperator*{\argmax}{argmax}

\newcommand*{\argmaxl}{\argmax\limits}

\allowdisplaybreaks
\usepackage{array}
\usepackage{arydshln} 
\usepackage{hyperref}

\begin{document}

\title{Articulated Clinician Detection Using 3D Pictorial Structures on RGB-D Data}

\author{Abdolrahim Kadkhodamohammadi, 
Afshin Gangi, 
Michel de Mathelin 
and Nicolas Padoy

\thanks{Abdolrahim Kadkhodamohammadi, Michel de Mathelin and Nicolas Padoy are with ICube, University of Strasbourg, CNRS, IHU Strasbourg, France ({E-mail}: kadkhodamohammad@unistra.fr).

Afshin Gangi is with the Radiology Department, University Hospital of Strasbourg, France, and also with ICube, University of Strasbourg, CNRS, France.
}}%

\markboth{}
{}
\maketitle

\begin{abstract}
Reliable human pose estimation (HPE) is essential to many clinical applications, such as surgical workflow analysis, radiation safety monitoring and human-robot cooperation. 
Proposed methods for the operating room (OR) rely either on foreground estimation using a multi-camera system, which is a challenge in real ORs due to color similarities and frequent illumination changes, or on wearable sensors or markers, which are invasive and therefore difficult to introduce in the room.
Instead, we propose a novel approach based on Pictorial Structures (PS) and on RGB-D data, which can be easily deployed in \textit{real} ORs. We extend the PS framework in two ways. First, we build robust and discriminative part detectors using both color and depth images. We also present a novel descriptor for depth images, called histogram of depth differences (HDD). Second, we extend PS to 3D by proposing 3D pairwise constraints and a new method that makes exact inference tractable.
Our approach is evaluated for pose estimation and clinician detection on a challenging RGB-D dataset recorded in a busy operating room during \textit{live} surgeries. We conduct series of experiments to study the different part detectors in conjunction with the various 2D or 3D pairwise constraints. Our comparisons demonstrate that 3D PS with RGB-D part detectors significantly improves the results in a visually challenging operating environment.

\end{abstract}

\begin{IEEEkeywords}
Medical computer vision, surgical activity analysis, clinician pose estimation, 3D pictorial structures, RGB-D Data
\end{IEEEkeywords}


\section{Introduction}\label{sec_introduction}
\IEEEPARstart{C}{}linicians and surgical staff are the main actors in the operating room (OR). Detecting clinicians and staff as well as localizing their body parts are therefore important to a wide range of applications, such as modeling and recognition of medical activities~\cite{agarwal_MNA2007,bardram_percom2011,lea_AMIA2013,padoy_voec2009,twinanda_ijcar2015},  performance assessment of the surgical team~\cite{gentric_JN2013}, safe human-robot interaction and collision avoidance~\cite{beyl_ijcar2015,ladikos_miccai2008}, and radiation safety monitoring~\cite{ladikos_MICCAI2010,loyrodas_ijcars2015}. 
Current methods proposed to estimate the poses of humans in the OR use multi-view systems and rely on foreground estimation~\cite{beyl_ijcar2015,ladikos_MICCAI2010}. But the challenge of obtaining a reliable foreground estimation in real operating rooms, which are cluttered, contain a lot of similar colors between equipment and loose clothes and are subject to illumination changes, impedes the deployment of such systems in real environments. Other methods that have been proposed to detect clinicians rely on body-worn sensors~\cite{agarwal_MNA2007,bardram_percom2011}. Such systems are however invasive and difficult to install in the OR.

\begin{figure}[tb]
\centering
\setlength{\tabcolsep}{1pt}
\renewcommand{\arraystretch}{0.5}
\begin{tabular}{c c}
\includegraphics[width=0.49\columnwidth]{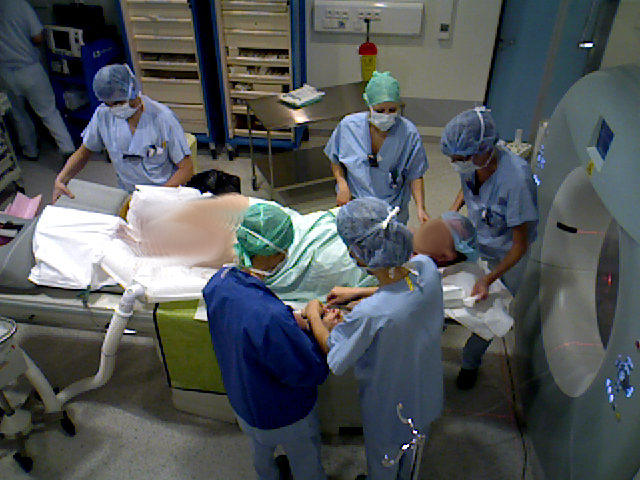} & 
\includegraphics[width=0.49\columnwidth]{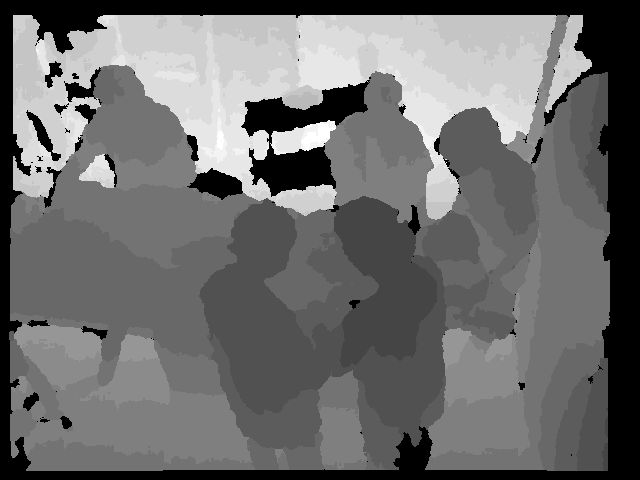}
\end{tabular}
\caption{Sample color and depth images captured using an affordable RGB-D camera.}
\label{fig:sampleImages}
\end{figure}

In this paper, we propose a novel vision-based approach to address clinician detection and pose estimation in {\it real} ORs. This approach relies on both color and depth images captured simultaneously by an RGB-D camera to cope with the aforementioned challenges of the operating environment. Figure~\ref{fig:sampleImages} shows an example of color image along with its corresponding depth image captured by an RGB-D camera. Each pixel in the depth image encodes the distance of the object surface to the camera. These distances are computed by decoding the deformation of a known pattern that is projected onto a scene using infrared light. Thus, the depth image is not sensitive to illumination changes, lack of texture and color similarities among different surfaces. In a cluttered OR, the depth image is also very useful to distinguish between two different 3D surfaces that appear visually similar and close to each other in the projected 2D image.

Human pose estimation (HPE) is a fundamental problem in the field of computer vision and has been studied for decades. In general, the proposed approaches for HPE can be divided into two main groups: holistic and part-based. On the one hand, holistic approaches localize human body parts by directly mapping image features into part locations~\cite{shotton_PAMI2012,toshev_CVPR2014}. But, to learn a direct mapping of image features into body poses, holistic approaches need a very large annotated training set. This data is not easy to collect and, to the best of our knowledge, no such dataset is available for the OR setting. On the other hand, part-based approaches estimate human poses using a set of part detectors combined with a deformation model that enforces inter-part displacement constraints. These approaches need a much smaller training set. We therefore base our approach upon the pictorial structures (PS) framework~\cite{felzenszwalb_IJCV2005}, which is the dominant part-based approach for HPE. It relies on color-based body part appearance models and pairwise deformation models between parts that are parametrized by their 2D pixel locations and orientations.  

In order to deal with the inherent challenges of real ORs, our proposed approach extends pictorial structures in two ways: by constructing robust appearance models using both color and depth images, and by using 3D deformation models along with an inference algorithm to perform exact and efficient inference. We base our work on a modified version of PS called flexible mixtures of parts (FMP), which is used for both human pose estimation and detection~\cite{yang_PAMI2013}. The FMP approach uses multiple mixtures in the PS framework to capture different part appearances and their co-occurrences. The approach also learns all model parameters with a structured SVM solver. However, it still relies on color-based appearance models and on 2D deformation models.

The most common descriptor on intensity images is the histogram of oriented gradient (HOG), which is referred to as I-HOG in this situation. In practice, color images might not always carry enough descriptive information in a visually complex environment such as the OR. We therefore construct robust and discriminative appearance models by combining both color and depth information. We also present a new descriptor for depth images, named {\it histogram of depth differences} (HDD), that encodes depth level changes. It uses small convolutional kernels for efficiency and provides a representation that describes each surface with respect to its neighbors. We compare the novel descriptor with two other descriptors for depth images: the HOG descriptor that is applied on depth images for comparison (D-HOG), and the histogram of oriented normal vectors (HONV)~\cite{tang_ACCV2012} that has been originally proposed for object detection in depth images.  

\begin{figure}[t]
\begin{center}
\begin{tabular}{ccc}

\includegraphics[width=0.31\linewidth]{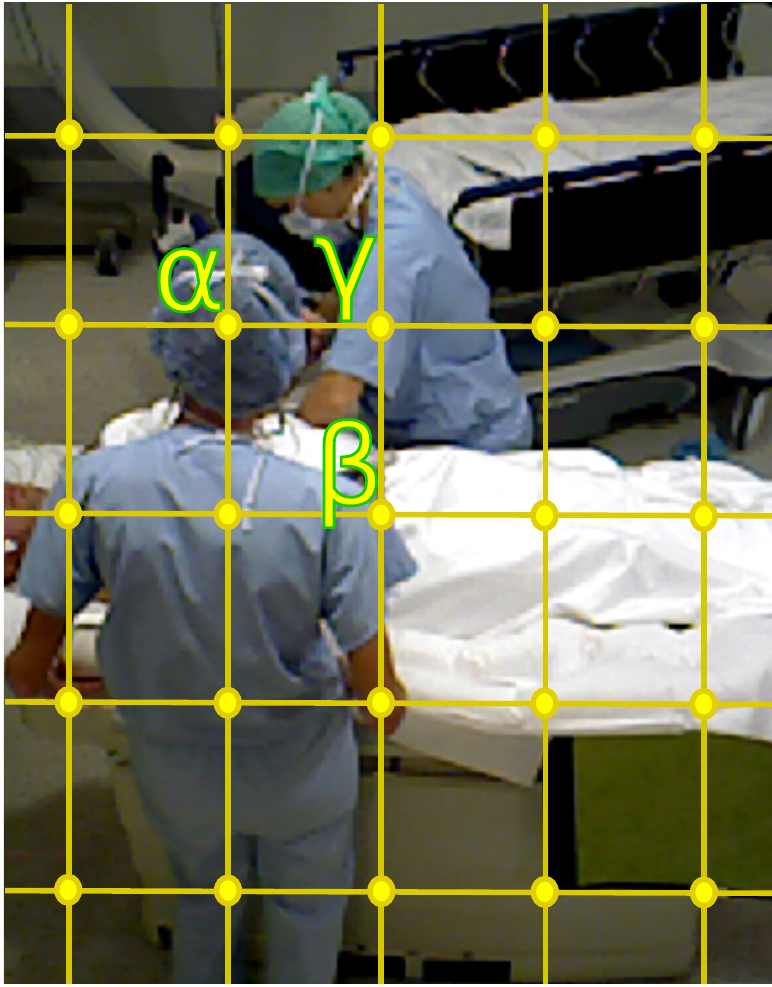} &
\includegraphics[width=0.31\linewidth]{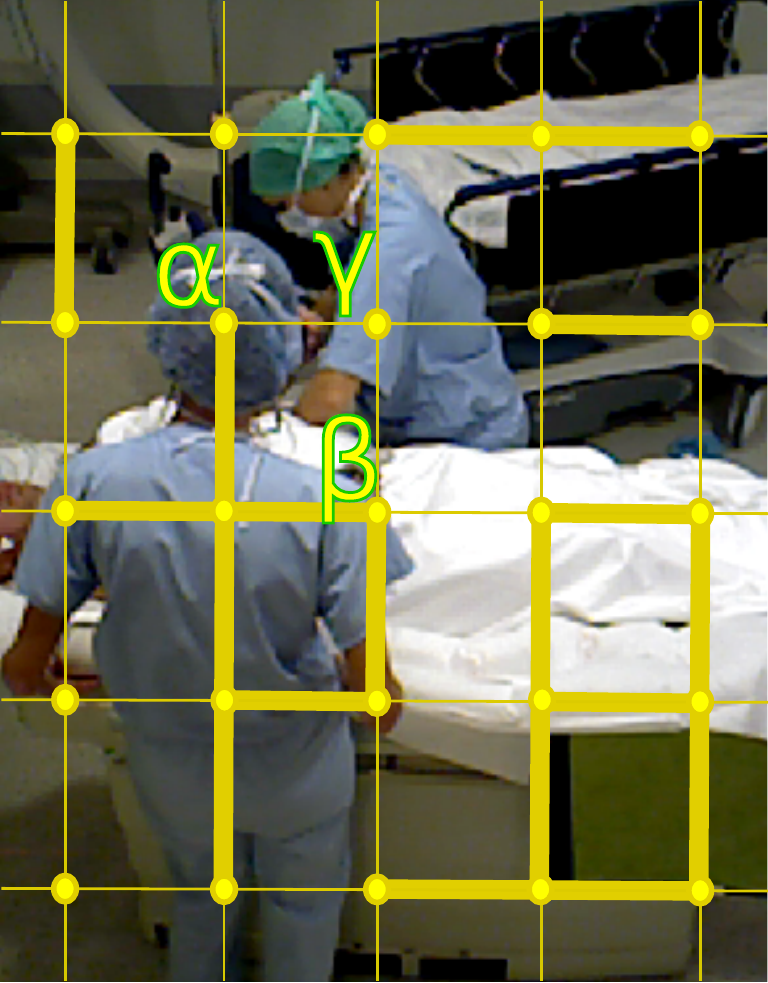} &
\includegraphics[width=0.31\linewidth]{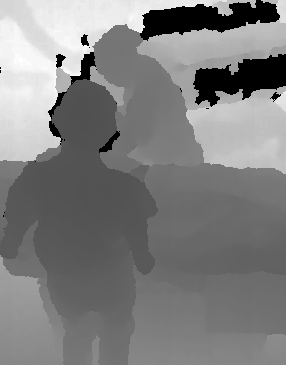} \tabularnewline
{\scriptsize (a)} &  {\scriptsize (b)} & \scriptsize{(c)}
\end{tabular}
\end{center}
   \caption{Two different connectivity maps for the same state space. Circles show nodes in the state space and edge thickness indicates the connectivity strength between two nodes: (a) connectivity map built using 2D pixel distances (b) the same connectivity map when real 3D positions of the nodes are considered (c) corresponding depth map used to re-project points into 3D. {\it Note: not all edges are represented in this picture.}}
   
\label{fig:DTIdea}
\end{figure}

In general, in part-based methods, the appearance models are used to obtain confidence scores for every part at every position in the image. 
The set that defines all possible positions for the parts is called state space. The deformation models are then used to weight connections between pairs of parts in the state space. The pose is finally estimated by an inference algorithm that propagates the scores according to the connectivity map. Figure~\ref{fig:DTIdea}(a) shows an example of state space. Each circle in the image represents one state, also called node. The thickness of an edge shows the connectivity strength between two states, here estimated using 2D pixel distances. However, by relying on 2D pixel distances, a path that is connecting two nodes lying over the same person ($\alpha$-$\beta$) can have a weight inferior to a path connecting two nodes lying over two different persons ($\alpha$-$\gamma$). Thus, inference approaches that are relying on such a connectivity map are mixing part detections of persons who appear close to each other in a 2D image but are not necessarily nearby in 3D. Moreover, in case of weak part detections, which is common in cluttered scenes such as the OR, inference can be additionally confused by false detections on the background. 

In this paper, we address this limitation by proposing an approach based on connectivity maps that rely on the true 3D positions of the nodes. To this end, we use the depth map to recover the 3D positions. Figure~\ref{fig:DTIdea}(b) illustrates the connectivity map for the same state space as in Figure~\ref{fig:DTIdea}(a), but where 3D positions are taken into account. One can notice that the nodes lying on the same person are strongly connected ($\alpha$-$\beta$), while connections crossing person boundaries are weak ($\alpha$-$\gamma$). 
Hence, by properly weighting the connections in 3D, message propagation across persons or across a person and the background is discouraged.

In terms of inference, Felzenszwalb and Huttenlocher~\cite{felzenszwalb_TR2004,felzenszwalb_IJCV2005} proposed an efficient approach for pictorial structures with tree-structured pairwise graphical models using the generalized distance transform (GDT) algorithm. This approach is exact, has a complexity linear in the number of states and can be used in 3D. 
However, it requires a fully connected regular 3D grid as state space, which makes the subsequent step, namely dynamic programming, intractable in terms of memory. With $n$ parts and a 3D state space of size $s_{3D}$, the memory complexity of dynamic programming is $O(n*s_{3D})$. 
Consequently, a very coarse discretization of the 3D state space would be required to make the approach tractable\footnote{
By using a 3D grid of $(200,200,200)$, more than 28GB of memory is needed to process one image. Due to the projection process, such a 3D grid is still not fine enough to cover all the image: 10\% of the appearance information is not taken into account.}, which would degenerate the performance.
Instead, we propose to use an irregular 3D state space along with the standard dynamic programming approach, which has quadratic complexity. By using 3D information, we can significantly reduce the number of connecting edges in the state space while retaining exact inference by excluding connections between nodes that are unreasonably far apart according to human body kinematic constraints. Although inference still has quadratic complexity, this reduction has a huge impact on run-time. Furthermore, the reduced map contains as many 3D nodes as 2D locations where the detector is evaluated. As a result, no appearance information is lost.

\begin{figure}[t]
\setlength{\tabcolsep}{1pt} 
\renewcommand{\arraystretch}{0.5} 

\begin{tabular}{ccc}
\includegraphics[width=0.32\columnwidth]{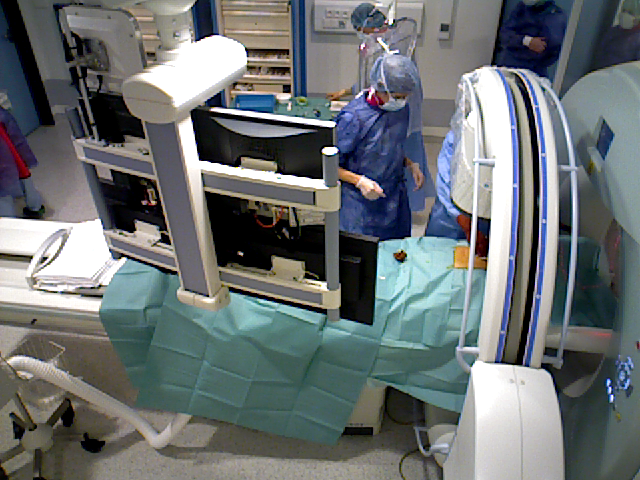} &
\includegraphics[width=0.32\columnwidth]{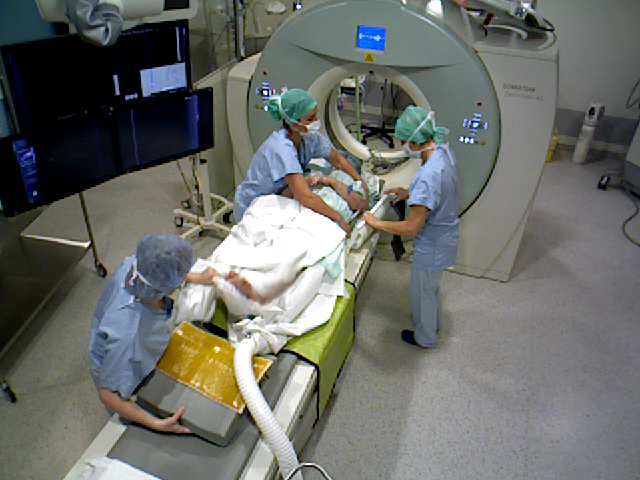} &
\includegraphics[width=0.32\columnwidth]{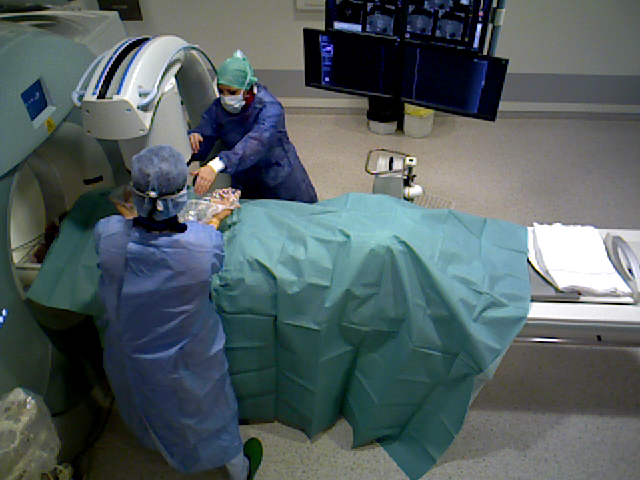} 
\end{tabular}
\caption{Sample images of the clinician pose dataset recorded from three different view points.}
\label{fig:dataset}
\end{figure}

In order to evaluate our approach, we have generated a new RGB-D dataset recorded in an operating room during live surgeries for seven half-days. Figure~\ref{fig:dataset} shows sample images from this dataset.  We have manually generated two types of annotations: upper-body bounding boxes of all clinical staff present in the scene to evaluate human detections and upper-body poses of clinical staff that have at least half of their upper-body parts visible to assess human pose estimation.  

The contributions of the paper can be summarized as follows: (1) we  extend the appearance model in pictorial structures to use both color and depth images; (2) we propose a novel feature descriptor for depth images; (3) we extend the deformation model in PS to use more reliable 3D constraints;
(4) we present an efficient algorithm to reduce the size of the 3D state space in order to make exact inference tractable; 
 (5) a new clinician pose dataset is generated;  
(6) different appearance models in combination with different deformation models are evaluated and compared with state-of-the-art methods for both the tasks of clinician pose estimation and detection. 

\section{Related work}

We hereby divide related work into two categories: first, we discuss methods that are addressing the fundamental problem of HPE using only vision-based approaches without focusing on a specific application; second, we discuss clinician detection methods proposed for the OR, which are either vision-based or sensor-based.

\subsection{Vision-based methods}

There is a significant body of work on human pose estimation and detection in the computer vision literature. 
In recent years, part-based approaches have been successfully applied on challenging datasets~\cite{kiefel_ECCV2014,yang_PAMI2013}. After pictorial structures was made tractable with exact inference by the seminal work of~\cite{felzenszwalb_IJCV2005}, it became the dominant part-based approach.

Appearance models in PS have been improved in different ways, for instance by estimating inter-part appearance dependencies~\cite{eichner_bmvc2009}, parsing the image iteratively for better visual features~\cite{ramanan_nips2006}, using deformable part shape models~\cite{zuffi_cvpr2012} and estimating common appearance models on test images~\cite{eichner_ACCV2012}. Yang and Ramanan~\cite{yang_PAMI2013} also proposed the FMP approach that extends PS by learning multiple mixtures for each part to capture appearance changes. Even though the appearance models have been greatly improved, they only rely on features computed on intensity images, which are not always discriminative in visually challenging environments. Moreover, two typical problems in HPE, namely background/foreground confusion in a cluttered scene and  inter-person part mixing in multi-person scenarios, are still unsolved, especially due to the limitations caused by the use of 2D spatial pairwise constraints and of tree-structured dependency graphs.

To mitigate the unreliability of 2D deformation models in PS, many approaches were proposed that rely on loopy dependencies between parts, such as temporal consistency~\cite{kadkhoda_IPCAI2014}, global consistency across an entire video~\cite{tokola_ICCV2013}, higher-order dependency using fields of parts~\cite{kiefel_ECCV2014} and two-stage PS with a loopy dependency graph to discourage part double counting~\cite{andriluka_IJCV2012}. Color, contour and optical flow have also been incorporated in stretchable models~\cite{sapp_CVPR2011}, which use  pairwise relationships to model body kinematic constraints and dependencies between the same parts in consecutive frames. In general, the loopy dependencies between parts make the exact inference intractable and the pairwise terms enforcing visual consistency can be confused, especially in cluttered scenes.

Part-based approaches have also been proposed for 3D human pose estimation in multi-view setups. They rely on 2D detectors in different ways, \eg by generating a set of body joint proposals that are refined using 3D reasoning~\cite{amin_BMVC2013,belagiannis_CVPR2014,sigal_IJCV2011} or by using 3D consistency to boost 2D pose estimation~\cite{amin_PR2014}. Burenius et al.~\cite{burenius_CVPR2013} proposed a 3D PS method for multiple views that performs exact inference. However, to cope with the time complexity of the inference and the high memory requirements of dynamic programming in a large 3D state space, a coarse discretization of the state space is used along with simple binary pairwise terms. Hofmann and Gavrila~\cite{hofmann_IJCV2011} introduced an exemplar-based multi-view 3D human pose estimation approach that generates 3D pose hypotheses per view by matching segmented foregrounds with a set of precomputed 2D exemplars. Pose trajectories  constructed using temporal information serve to generate 3D texture models. Then, the pose hypotheses are projected to all views and multi-view likelihood scores are computed based on shape and texture consistencies to predict the 3D poses. Both the 3D candidate generation and the multi-view likelihood computation are relying on foreground estimation, which is challenging to obtain in a complex environment like the OR.

In addition, holistic approaches have also been proposed to predict the body joints from the image features. Recently, promising approaches based on deep convolutional neural networks (CNNs) have been proposed, such as cascaded CNNs~\cite{toshev_CVPR2014} and cascaded CNNs combined with position refinement model to fine tune the estimated joint positions~\cite{tompson_CoRR2014}. However, due to the use of CNNs with huge numbers of parameters, very large training sets are required. 
Haque et al.~\cite{haque_arXiv2016} proposed a 3D human pose estimation on a depth image based on a shallow CNN and a deep convolutional and recurrent neural network to jointly predict body part occlusions and regress for body joint positions. A shallow convolutional network is used to provide a viewpoint-invariant feature map. These features are then fed into the deep convolutional network with recurrent connections to iteratively predict body part occlusion masks and correct the previously predicted poses. However, this approach is only evaluated for single person scenarios captured in a controlled laboratory environment.

With the availability of the affordable RGB-D sensors, Shotton et al.~\cite{shotton_PAMI2012} proposed a very promising holistic approach for HPE applicable in scenarios where a foreground mask can be obtained. The approach uses a random forest on depth images alone. A commercial and extended implementation of this approach is provided with the Microsoft Kinect RGB-D camera. We have qualitatively evaluated this implementation in an OR: it often fails at detecting the clinicians and mixes the body parts of multiple persons\footnote{Qualitative results are presented in the supplementary video. Quantitative comparison is not possible because the software cannot be run offline on recorded videos.}. We believe this low performance is due to two reasons: (1) the cluttered scene makes the foreground estimation fail; (2) the random forest has not been trained for the loose clothes worn by the clinicians and for the top view of the camera, which has to be mounted on the ceiling. Retraining the forest is however a challenging task, requiring a very large dataset difficult to generate. In~\cite{ye_ICCV2011,baak_ICCV2011}, the 3D point cloud reconstructed from depth data is used to estimate 3D human poses with exemplar-based approaches. In \cite{liu_ICIP2013,jafari_ICRA2014}, color and depth features are used along with temporal tracking to detect persons. These approaches however all require foreground segmentations. They are therefore challenging to apply in the cluttered and busy OR environment.


\subsection{Clinician detection methods for the OR}

Recently, Ladikos et al.~\cite{ladikos_MICCAI2010} have introduced a 16 camera multi-view system to compute and show the accumulation of the radiation risk per body part. The body parts are localized based on background subtraction and shape from silhouette. In~\cite{beyl_ijcar2015}, a system for safe human robot interaction is proposed, which uses four Kinect cameras to track clinicians. The middleware NiTE~\cite{openni}, which uses a method similar to~\cite{shotton_PAMI2012}, is used to obtain body skeletons per view. Then, the skeletons are fused in 3D. However, both \cite{beyl_ijcar2015} and \cite{ladikos_MICCAI2010} have been evaluated in single person scenarios recorded in controlled laboratory environments. Moreover, these approaches require a foreground estimation that is difficult to compute in real ORs. In \cite{agarwal_MNA2007}, radio frequency identification (RFID) tags are used in a workflow analysis system to detect staff that are present in the OR. This system can only detect the presence of the staff and is not able to localize them. Body-worn sensors are used in~\cite{bardram_percom2011} to track clinicians in the OR. Due to the complexity of the system setup, it is only tested during a surgical simulation that took place in a laboratory rigged to resemble an OR.

In \cite{kadkhoda_IPCAI2014}, a method is proposed to estimate 3D human poses in the OR using discrete optimization over RGB-D sequences. A publicly available body part detector is used, which is based on a random forest and has not been trained for the OR environment. An energy optimization over a graph defined across an entire sequence is used to enforce temporal consistency and also to cope with the detection failures. To perform the optimization, ground-truth positions in the first and last frames are needed. Due to these requirements, the approach is not suitable for real-time applications. The approach has also not been evaluated on a dataset recorded during real surgeries. Furthermore, the high failure of the detector indicates the need for a body-part detector trained specifically for the OR. 

In~\cite{kadkhoda_miccai2015}, we present a preliminary version of this work that is hereby extended in the following ways. First, a new deformation model is defined, which takes benefit of 3D distance metrics. Second, an efficient algorithm is presented to make exact 3D inference tractable via state space pruning. Third, to quantitatively evaluate human detection, the dataset presented in~\cite{kadkhoda_miccai2015} is augmented with upper-body bounding boxes of all clinical staff appearing in the views. Fourth, additional experiments are presented to quantitatively evaluate and compare the approach with state-of-the-art methods for both human detection and pose estimation.

\section{Method}
In this section, we briefly describe the flexible mixtures of parts approach~\cite{yang_PAMI2013}, which serves as a base to develop our method for clinician pose estimation and detection. We then present our RGB-D based appearance model followed by our proposed 3D deformation model. Finally, an inference algorithm is presented, which enables exact and tractable inference in PS with 3D pairwise constraints.

\subsection{Flexible mixtures of parts}
\label{sec:FMP}
The FMP approach represents human poses using a set of body parts~\cite{yang_PAMI2013}. The pose estimation is broken down into a set of 2D part detections combined with pairwise constraints between parts. Formally, the problem is represented as an energy function defined over a tree-structured graph $G=(V,E)$ whose nodes are the body parts and whose edges are the pairwise kinematic constraints. Let $I$ be the input intensity image. The state of a body part $i$ is specified by its location $l_i$ and its part type $t_i$. Part types are used to learn multiple part detectors and capture appearance changes. The energy function $S(.,.,.)$ is defined as: 
\begin{dmath}\label{eq:Energy2D}
S(I,l,t) = \sum_{i\in V}w_{i}^{t_i}. \rho(I,l_i) + \sum_{ij\in E} w_{ij}^{t_i,t_j}.\psi_{2D}(l_i-l_j) + \sum_{i\in V} b_i^{t_i} + \sum_{ij\in E} b_{ij}^{t_i,t_j},
\end{dmath}
where in a model with $N_p$ parts, $l=(l_1,...,l_{N_p})$ indicates the pose of the person using the positions of all body parts. The first term is the part appearance model and the second term is the deformation model that enforces pairwise constraints. The two last terms are capturing part compatibility, where $b_i^{t_i}$ is the score of choosing a particular mixture type for part $i$ and  $b_{ij}^{t_i,t_j}$ encodes the co-occurrence probability of part types. 

\noindent\textbf{Appearance model.} The appearance model, also referred to as part detector, computes the score of placing a body part at image location $l_i$ using a part template $w_i$ that is learned during training and a feature vector $\rho(I,l_i)$. The feature vector $\rho(I,l_i)$ is extracted at image location $l_i$. In FMP~\cite{yang_PAMI2013}, HOG on color image is used.

\noindent\textbf{Deformation model.} The spatial pairwise constraints are enforced by $w_{ij}$ and $\psi_{2D}(l_i-l_j)$. The weights $w_{ij}$ encode the deformations between pairs of parts and are learned during training. $\psi_{2D}(l_i-l_j)=[\overline{dc},{\overline{dc}}^2, \overline{dr}, {\overline{dr}}^2]^T$ captures the relative displacement of part $i$ w.r.t. part $j$, where $[\overline{dc},\overline{dr}]=[dc,dr]-[ac_{ij},ar_{ij}]$, $dc$ and $dr$ are the displacements along the columns and rows of the image, and $ac_{ij}$ and $ar_{ij}$ are the average kinematic distances estimated during training between these two  parts. Note that, by including $ac_{ij}$ and $ar_{ij}$, this notation for $\psi_{2D}$ is slightly different from the one in \cite{yang_PAMI2013} to allow for better comparison with the generalization to 3D given below.

\noindent\textbf{Learning and inference.} In a supervised learning paradigm, a structured prediction objective function can be defined to learn all parameters, including part templates, deformation parameters and co-occurrence relations. More details can be found in \cite{yang_PAMI2013}. Given input color image and the learned model parameters, inference corresponds to finding $(l^*,t^*) =\argmax_{l,t} S(I,l,t)$. This optimization can be solved efficiently and exactly for tree-structured and pixel-based pairwise dependencies using the generalized distance transform (GDT) and dynamic programming. 

\subsection{3D pictorial structures on RGB-D data}

Given the availability of synchronous and aligned color and depth images, it is natural to think of models that can benefit from the complementary information. This can be achieved in the PS framework by extending both the appearance and deformation models: 

\begin{dmath}\label{eq:Energy3D}
S(I,D,l,t) = \sum_{i\in V}w_{i}^{t_i}.\phi(I,D,l_i) + \sum_{ij\in E} w_{ij}^{t_i,t_j}.\psi_{3D}(D,l_i,l_j) + \sum_{i\in V} b_i^{t_i} + \sum_{ij\in E} b_{ij}^{t_i,t_j},
\end{dmath}
where $D$ is the aligned depth image and $\psi_{3D}$ models the 3D pairwise constraints. $\phi(I,D,l_i)$ is computed by concatenating the features extracted from the color and depth images. Following common practice in the literature, we use the HOG descriptor on intensity images (I-HOG). We compare three descriptors on depth images, namely D-HOG (HOG applied on the depth image), HONV~\cite{tang_ACCV2012} and HDD, defined below.  

\subsection{Histogram of depth differences (HDD)}

\begin{figure}[tb]
\renewcommand{\arraystretch}{1}
\centering
\begin{minipage}[b]{.22\columnwidth }
\centering
\begin{tabular}{|c|c|c|}
\hline
0  & 0 & 0 \\ \hline
-1 & \,0\, & \,1\, \\ \hline
0  & 0 & 0 \\ \hline
\end{tabular}
\end{minipage}
\begin{minipage}[b]{.22\columnwidth }
\centering
\begin{tabular}{|c|c|c|}
\hline
-1  & 0 & 0 \\ \hline
0 & \,0\, & \,0\, \\ \hline
0  & 0 & 1 \\ \hline
\end{tabular}
\end{minipage}
\begin{minipage}[b]{.22\columnwidth }
\centering
\begin{tabular}{|c|c|c|}
\hline
0  & -1 & 0 \\ \hline
\,0\, & \,0\, & \,0\, \\ \hline
0  & 1 & 0 \\ \hline
\end{tabular}
\end{minipage}
\begin{minipage}[b]{.22\columnwidth }
\centering
\begin{tabular}{|c|c|c|}
\hline
0  & 0 & -1 \\ \hline
\,0\, & \,0\, & \,0\, \\ \hline
1  & 0 & 0 \\ \hline
\end{tabular}
\end{minipage}
\caption{Four different kernels that capture local level changes in depth images.}
\label{fig:HDDKernel}
\end{figure}

A depth image encodes object surface distances w.r.t. the depth camera. We therefore propose a descriptor on depth images that extract features based on relative surface distance changes. A related idea has been explored earlier in~\cite{shotton_PAMI2012} that uses an ensemble of deep trees as body part detectors. The decision function of the trees relies on the relative surface distances. However, training such a deep forest requires a very large training set and it also requires retraining for each application. Instead, we introduce a simple yet efficient new descriptor that encodes local surface level changes. The descriptor uses 4 kernels that are shown in Figure~\ref{fig:HDDKernel}. 
Let $K_k$ be one of the HDD kernels with $k\in\{1,...,4\}$. The normalized convolution response at position $(x,y)$ is defined as
\begin{equation}
\label{eq:HDD}
C_{ks}(x,y) = (K_k \; * \; P_s(x,y))/ D_s(x,y),
\end{equation}
where $s\in\{1,...,n\}$ is the scale of the depth image. $P_s(x,y)$ and $D_s(x,y)$ are respectively the depth image patch and the depth value at location $(x,y)$ for scale $s$. The normalization of the response by the inverse of the depth value at the patch center guarantees that the feature is depth invariant. The convolution is also applied over a scale space to encode the changes in different spatial neighborhoods. To compute the descriptor, the convolution responses are quantized and the image is also divided into non-overlapping windows, called cells. The descriptor is then calculated per cell by computing a 3D histogram of kernel, scale and quantization levels.

\subsection{3D pairwise constraints}
In this section, we propose novel approaches to enforce the pairwise constraints in 3D. We define 
\begin{dmath}\label{eq:3D1pairwise}
\psi_{3D}^{1}(D,l_i,l_j) = [\overline{dx}, {\overline{dx}}^2, \overline{dy}, {\overline{dy}}^2, \overline{dz}, {\overline{dz}}^2]^T,  
\end{dmath}
where $ [\overline{dx}, \overline{dy}, \overline{dz}] = [dx, dy, dz] - [ax_{ij},ay_{ij},az_{ij}]$, $(ax_{ij},ay_{ij},az_{ij})$ are the average kinematic displacements in each direction between parts $i$ and $j$ estimated during training and $(dx, dy, dz)$ are the relative displacements between the two parts in x, y and z directions. $\psi_{3D}^{1}$ is a natural generalization of $\psi_{2D}$ to 3D. Since in 3D, body part lengths are relatively constant, we enforce these constraints explicitly by using the absolute 3D Euclidean distance between body joints. Let $\psi_{3D}^{2}$ and $\psi_{3D}^{3}$ be defined as 
\begin{dmath}\label{eq:3D2pairwise}
\psi_{3D}^{2}(D,l_i,l_j) = [|\overline{d_{3D}}|,\overline{dx}, {\overline{dx}}^2, \overline{dy}, {\overline{dy}}^2, \overline{dz}, {\overline{dz}}^2]^T,  
\end{dmath}  
\begin{dmath}\label{eq:3D3pairwise}
\psi_{3D}^{3}(D,l_i,l_j) = [|\overline{d_{3D}}|,\overline{dx},\overline{dy},\overline{dz}]^T,  
\end{dmath}  
where $\overline{d_{3D}}=\|[dx,dy,dz]\|-a_{ij}$ and $a_{ij}$ is the average 3D Euclidean distance between parts $i$ and $j$ estimated during training. $\psi_{3D}^{2}$ enforces body kinematic constraints by using not only relative displacement directions and magnitudes along the 3D axes, but also absolute 3D Euclidean distances between joints. In $\psi_{3D}^{3}$, we drop the square terms to only rely on absolute 3D Euclidean distances~$\left(\overline{d_{3D}}\right)$ for enforcing part lengths.

In practice, obtaining precise 3D positions is not always possible due to noise. This is common in datasets obtained from affordable RGB-D cameras and, in such cases, the 2D annotations are incorrectly re-projected back to 3D. During training, these incorrect 3D points can introduce a large error to the part lengths. We therefore present a pairwise model $\psi_{3D}^{4}$ that combines 2D and 3D constraints by relying both on absolute 3D Euclidean distance and on pixel displacement consistency. As will be shown in the experiments, incorporating the 2D distances will prevent the learning algorithm from being misled by noisy 3D positions. $\psi_{3D}^{4}$ is defined as
\begin{dmath}\label{eq:2.5Dpairwise}
\psi_{3D}^{4}(D,l_i,l_j) = [|d_{3D}|, \overline{dc},{\overline{dc}}^2,\overline{dr}, {\overline{dr}}^2]^T.  
\end{dmath}

\begin{algorithm}[t]
\caption{Construction of the state space's neighborhood map}
\label{alg:buildingNG}
\begin{algorithmic}[1]
\State $maxDist_{3D} \gets 0.9$ \Comment Distances are in meter
\State neighbors = $\emptyset$ \Comment 2D array to store the neighbourhood map
\For{ $i = 1$ to $L$ } \Comment{$L$: total number of the states}
\State neighbors[i] = $\emptyset$ \Comment Sorted array based on the distances
\State $C = getCandidates(x_i,depth[i])$\Comment{$depth$: array containing the depth values}
\For{ each $x_n \in C$}
\State $nodeDist = distance3D(x_i, x_n, depth)$\Comment Euclidean distance between $x_i$ and $x_n$
\If{$nodeDist < maxDist_{3D}$}
\State $insert(neighbors[i], (nodeDist,x_n)\; )$ 

\EndIf
\EndFor
\EndFor
\end{algorithmic}
\end{algorithm}

\noindent\textbf{Learning and inference.} The same approach as \cite{yang_PAMI2013} is used to automatically learn all parameters. 
To describe the inference, as in \cite{yang_PAMI2013}, we define $z_i=(l_i,t_i)$ and re-write the optimization as
\begin{dmath}
\label{eq:opt1}
(l^*,t^*) = \argmax_{z_i} \sum_{i\in V} f_i(I,D,z_i) + \sum_{ij \in E} d_{ij}(D,z_i, z_j),
\end{dmath}
where $f_i(I,D,z_i) = w_i^{t_i}.\phi(I,D,l_i)+ b_i^{t_i}$ and $d_{ij}(D,z_i, z_j) = w_{ij}^{t_i,t_j}.\psi_{3D}(D,l_i,l_j) + b_{ij}^{t_i, t_j}$. Using the tree-structured pairwise dependency the inference can be written as
\begin{dmath}
\setlength{\arraycolsep}{0.5pt} 
\renewcommand{\arraystretch}{0.5} 
\label{eq:msgPropagation}
\begin{cases}
\argmaxl_{z_{r}} \left(f_{r}\left(I,D,z_{r}\right) + \sum_{q\in child\left(r\right)} \mu_{z_q \rightarrow z_{r}}\right) \\
\begin{array}{ll}
\mu_{z_q \rightarrow z_p}=&\argmaxl_{z_q} \bigg(f_{q} (I,D,z_{q}) + d_{qp} (D, z_q, z_p ) \\ 
&\quad\quad\quad\;\;+\sum_{ch\in child (q)} \mu_{z_{ch} \rightarrow z_q}\bigg), 
\end{array}
\end{cases}
\end{dmath} 
where $r$ stands for the root part in the tree.

We start to propagate the scores from the leave nodes upward to the root part by using dynamic programming. Once all the scores have been propagated to the root part, pose confidence is available in the root and the corresponding full pose can be obtained by back-tracing the scores. Standard child-parent score propagation is quadratic in the size of the state space since every combination of child-parent nodes needs to be evaluated. Inference corresponds to the same optimization problem as in \cite{yang_PAMI2013}. However, the complete 3D grid required to apply the linear time GDT \cite{felzenszwalb_TR2004} makes dynamic programming intractable in memory, as mentioned in the introduction. In our approach, we therefore keep a small state space, which is generated by re-projecting the 2D nodes to 3D using the depth information. By using 3D information, we are also able to significantly reduce the neighborhood map of the state space. This is achieved by removing the connections between nodes that are too far apart in 3D. Even though this will only reduce the quadratic complexity of the inference by a constant value, it has a dramatic effect during the learning stage that uses the inference intensively. The neighborhood map can be constructed efficiently without comparing the distances between all states, as described in Algorithm~\ref{alg:buildingNG}. We define $maxDist_{3D}$ to be the maximum allowed part length in 3D. For a given $maxDist_{3D}$ and node $x_i$ at location $l_i$, only the 2D nodes at a 2D distance $maxDist_{2D}$ of $x_i$ need to be inspected, where 
\begin{equation}
\label{eq:patchSize}
maxDist_{2D} = maxDist_{3D}/\left(res\times depth(l_i)\right)
\end{equation}
and $res$ is the resolution of a camera pixel obtained from the intrinsic parameters. Since the nodes can be stored in a 2D array, these candidates can be accessed in constant time. As the 2D criteria cannot guaranty that the corresponding 3D nodes are within distance $maxDist_{3D}$, this condition is further checked among all potential candidates. A large distance of 0.9 meter is chosen for $maxDist_{3D}$, to make sure that the global optimum is not missed. Once the neighborhood map is constructed, it is used in (\ref{eq:msgPropagation}) to propagate the scores between all body parts.

\section{Experimental results}\label{sec:resDis}

We have generated a novel RGB-D dataset recorded in an operating room using an \textit{Asus Xtion Pro Live} camera. The camera position is changed among three possible locations to capture the room from different view points, shown in Figure~\ref{fig:dataset}. The dataset consists of 1451 annotated frames that are evenly selected across seven half-days of recordings, and 173 negative frames that do not contain any human for training. The dataset contains 3023 bounding boxes annotating all members of the clinical team. If the head or more than 50\% of the upper-body of a person is occluded, it is labeled with a {\it difficult} flag. 476 persons are annotated as {\it difficult} in the dataset. We also annotate the clinical staff with ground-truth positions for nine upper-body parts, namely head, left and right shoulders and hips, as well as left and right elbows and wrists. The pose annotation is only provided for staff who have more than six parts visible. We therefore obtain 1991 persons with pose annotations. The dataset covers many visual challenges, such as severe part foreshortening, clutter, occlusion and multi-person scenarios. We divide the dataset into seven disjoint sets where each set only contains frames that belong to the same half-day recording. A leave-one-out scheme is used during our experiments, so that one set is used as test set and the rest as training set. The average results of the seven-fold cross validation are reported during the evaluation.

We compute all the descriptors using the same parameters as in~\cite{yang_PAMI2013}, \ie the cell size is $6\times6$ pixels and the number of mixtures set to six. Also, we similarly normalize the descriptor responses using the L2-Hys normalization scheme, defined as a L2-norm where the maximum value is limited to 0.2. We compute the HDD descriptor in three scales and coarsely quantized the convolution responses into ten levels to be robust to noise and spatial distortions.  At test time, to be robust to body part scale changes in an image, we create an image pyramid by repeatedly smoothing and subsampling the image. We then evaluate our model on the image pyramid and perform non-maximum suppression to detect body parts of different sizes.
   
\subsection{Clinician pose estimation}

\begin{figure*}[t]
\setlength{\tabcolsep}{1pt} 
\renewcommand{\arraystretch}{0.5} 

\begin{tabular}{cccc}
\includegraphics[width=0.24\linewidth]{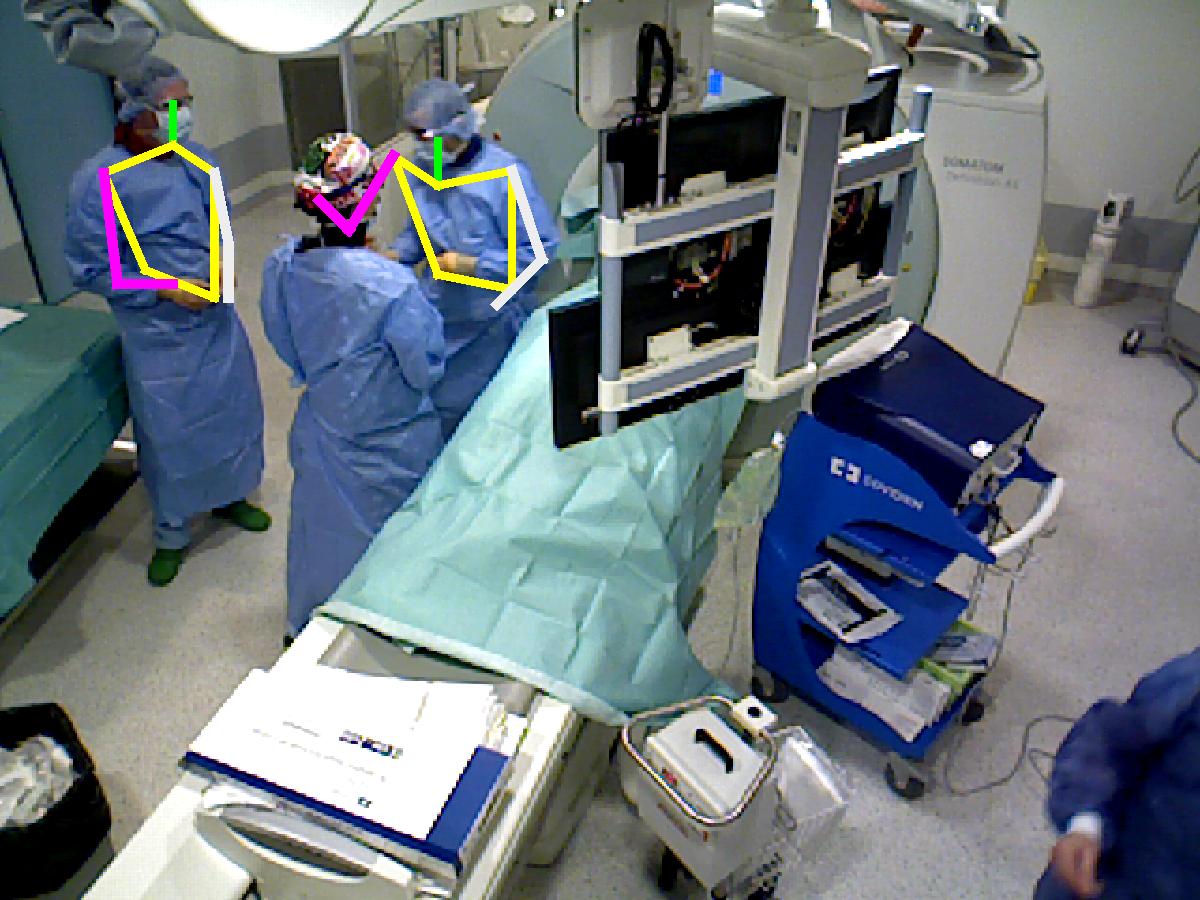} &
\includegraphics[width=0.24\linewidth]{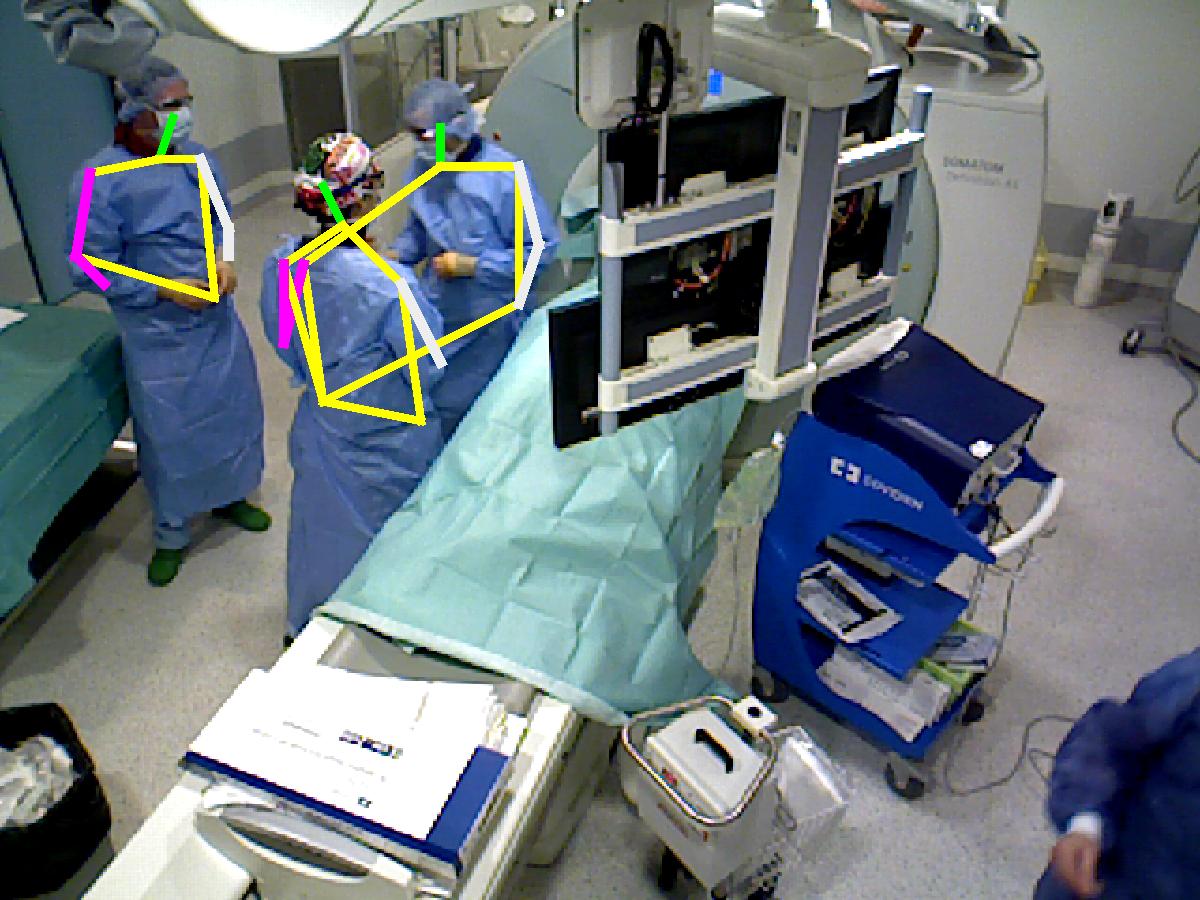}&
\includegraphics[width=0.24\linewidth]{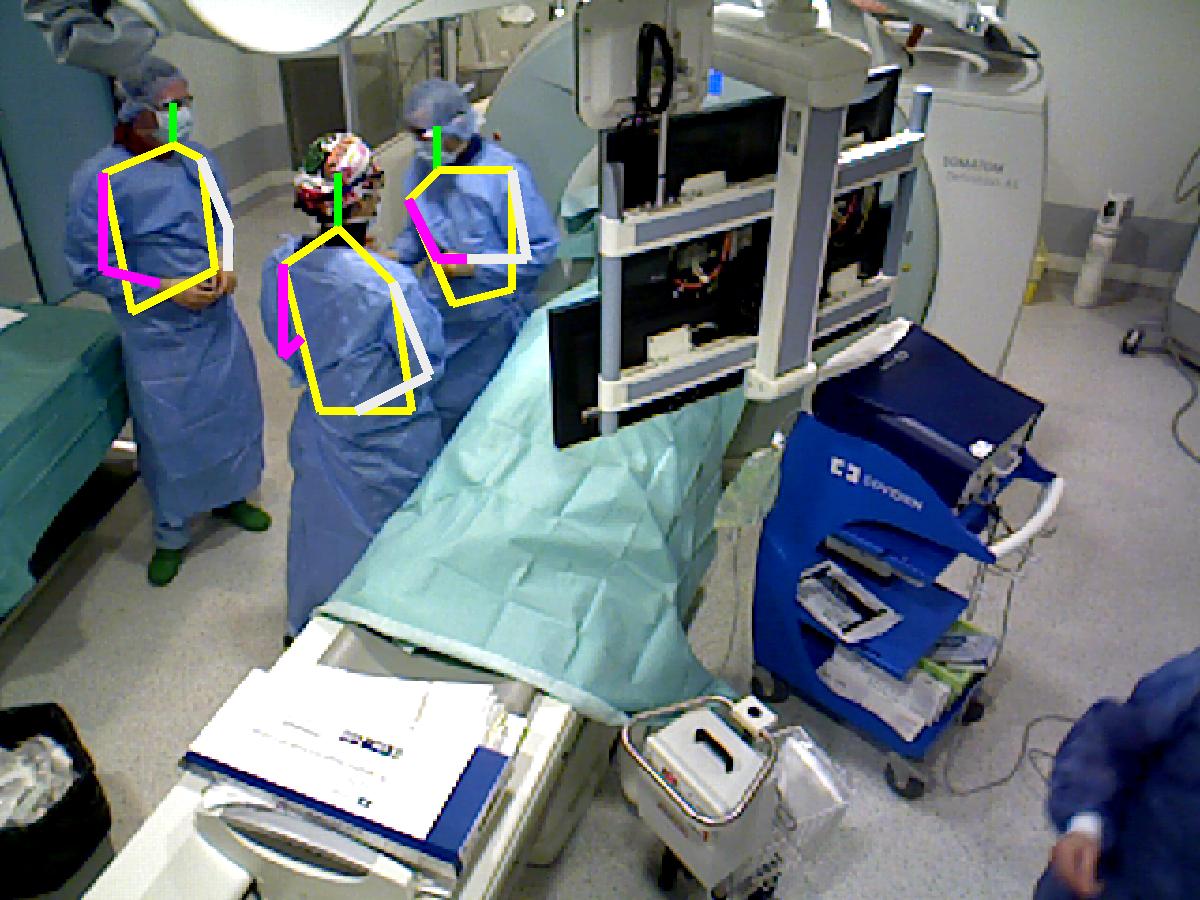}&
\includegraphics[width=0.24\linewidth]{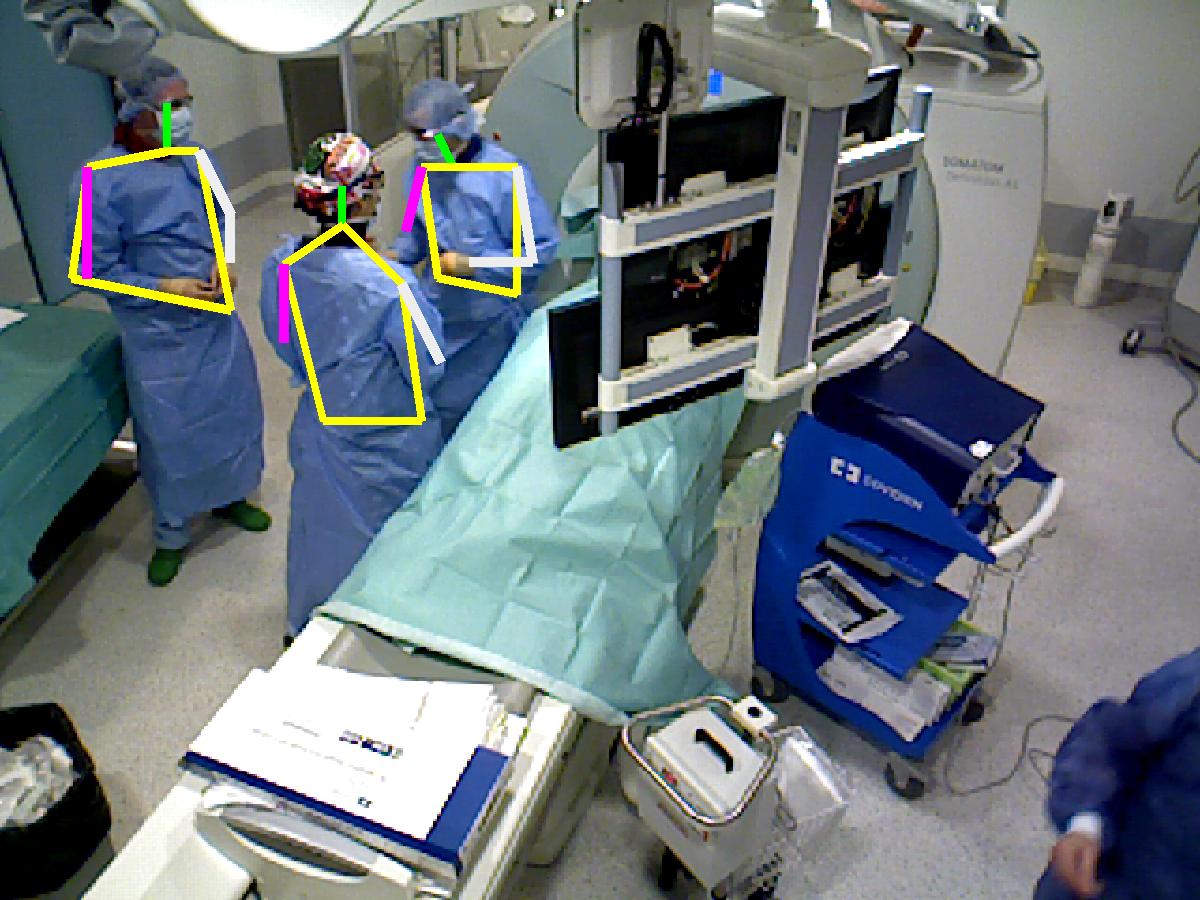}\\
\includegraphics[width=0.24\linewidth]{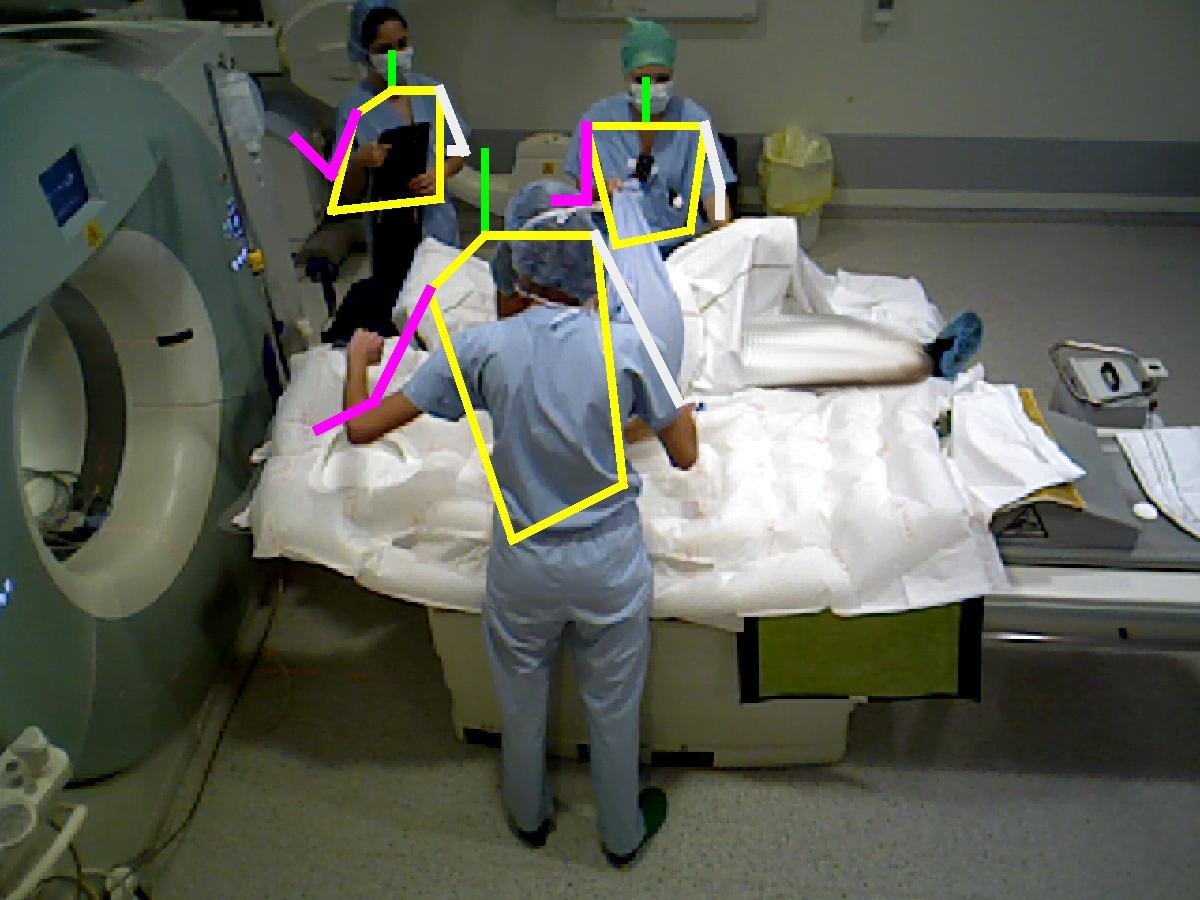}&
\includegraphics[width=0.24\linewidth]{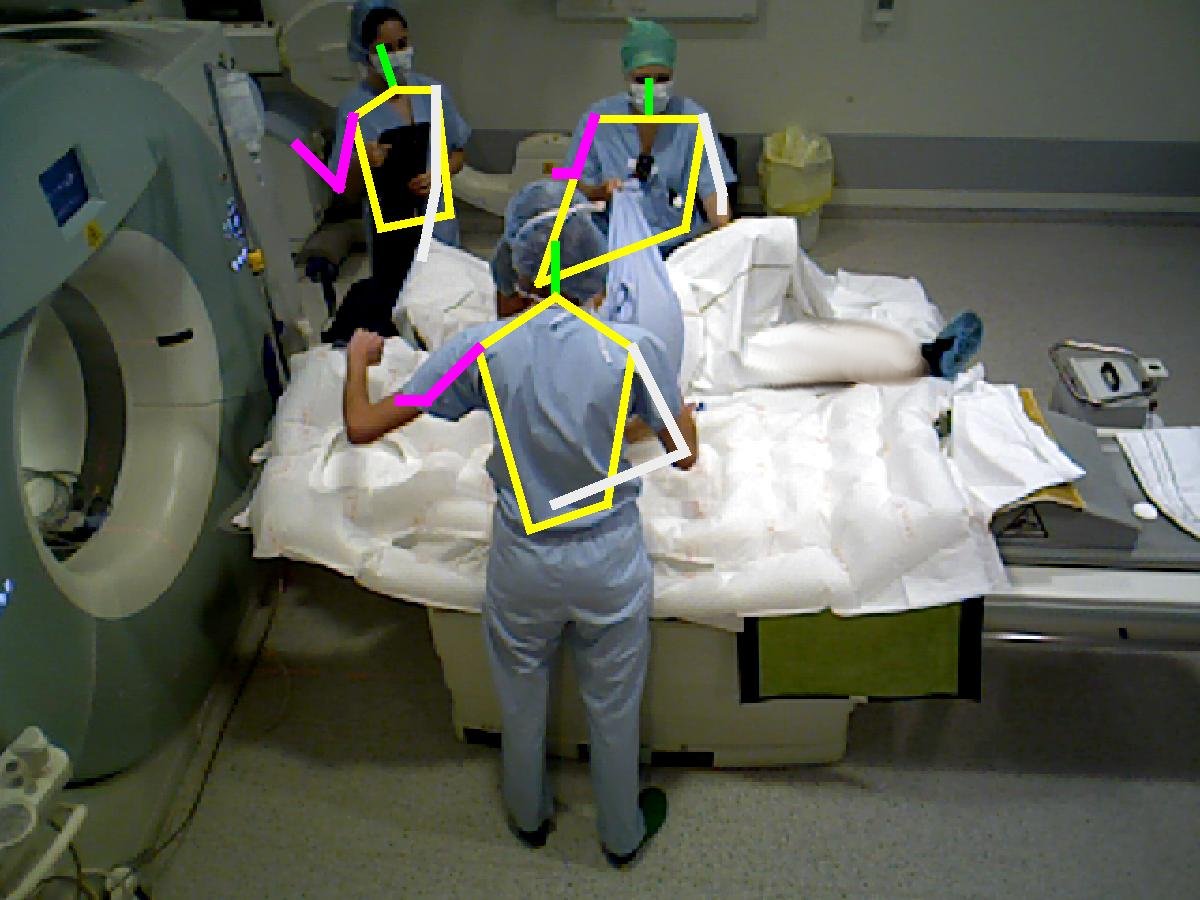}&
\includegraphics[width=0.24\linewidth]{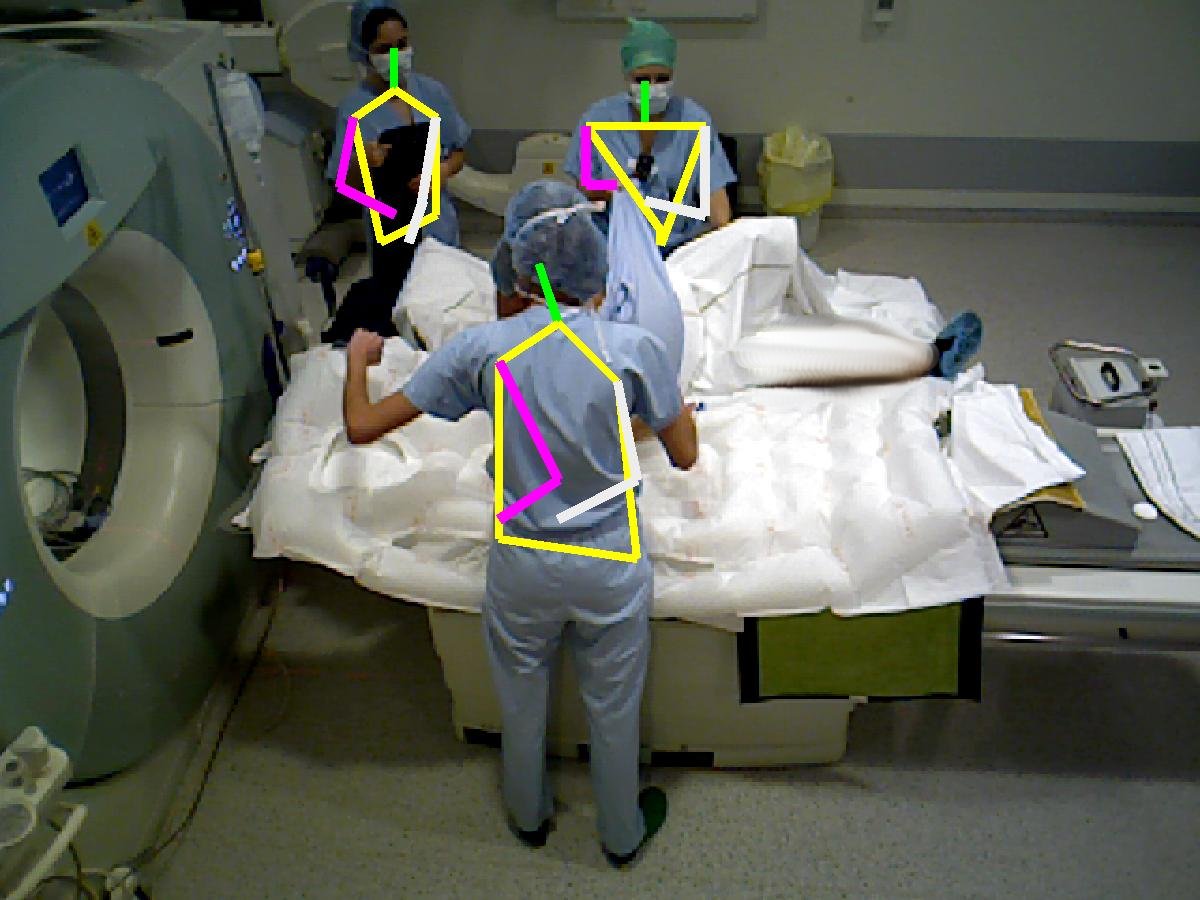}&
\includegraphics[width=0.24\linewidth]{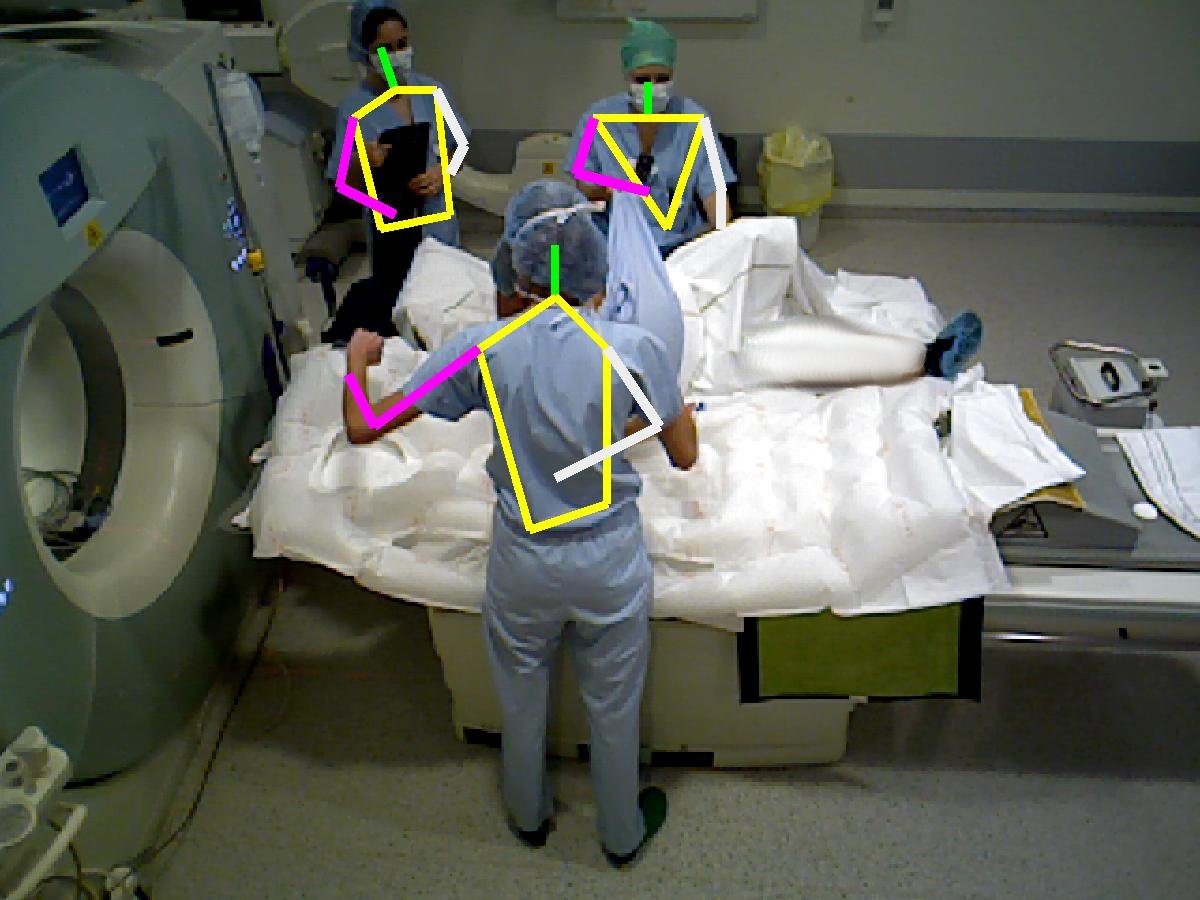}\\
\includegraphics[width=0.24\linewidth]{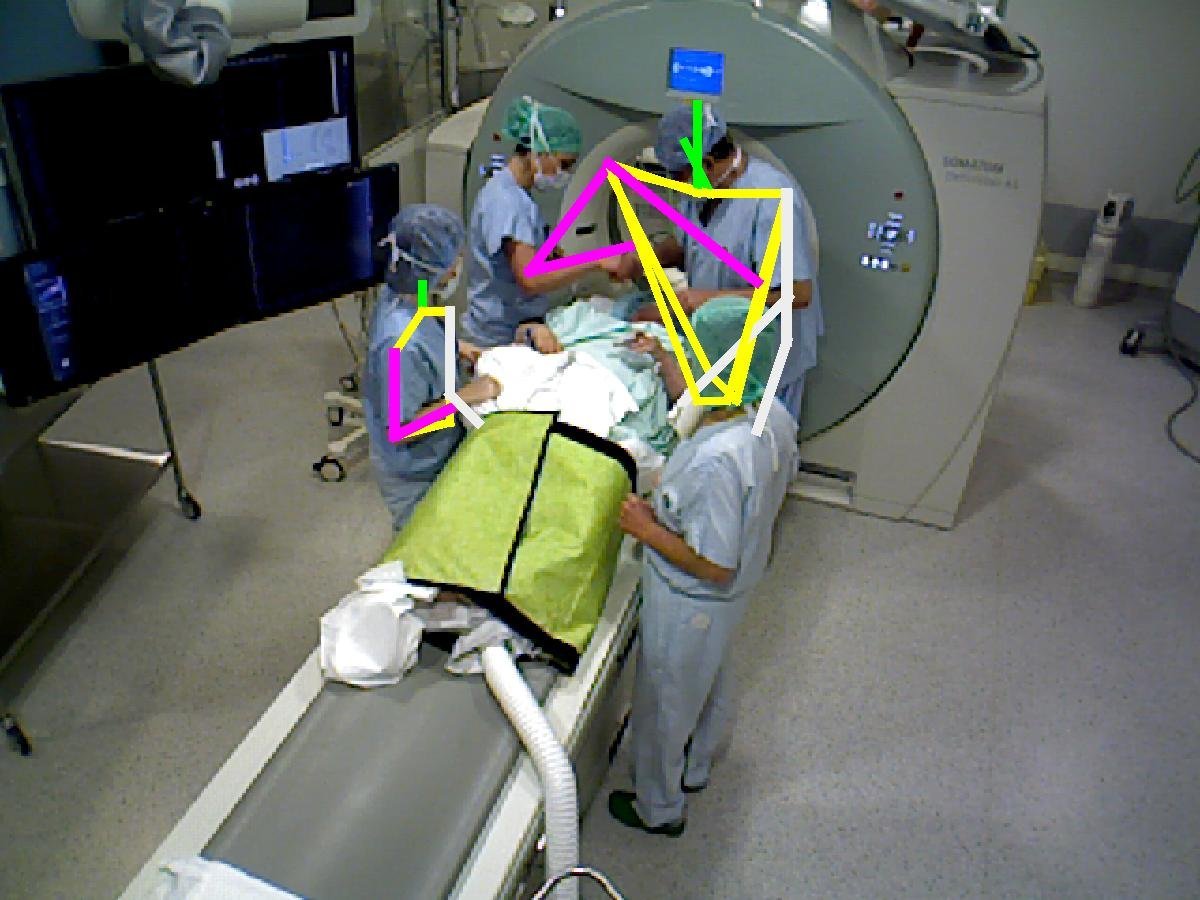}&
\includegraphics[width=0.24\linewidth]{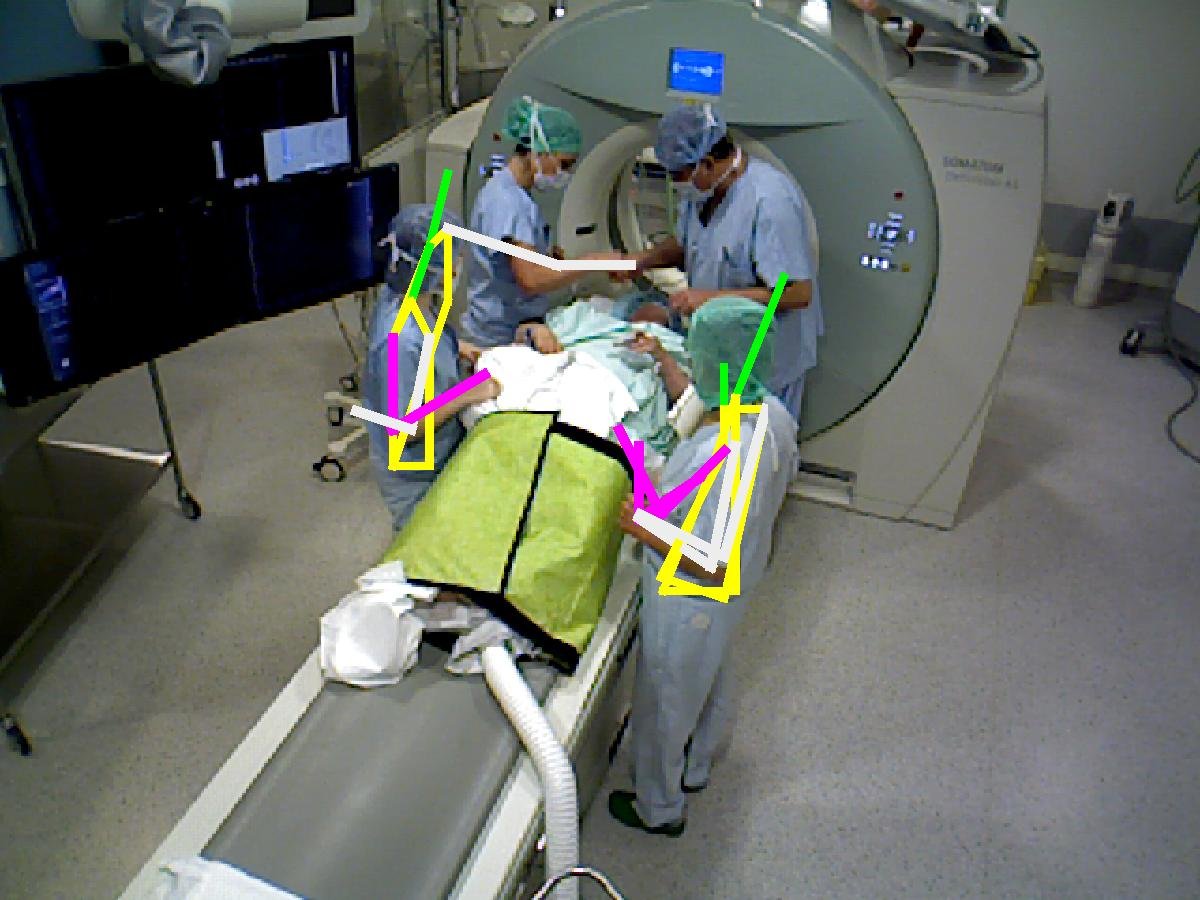}&
\includegraphics[width=0.24\linewidth]{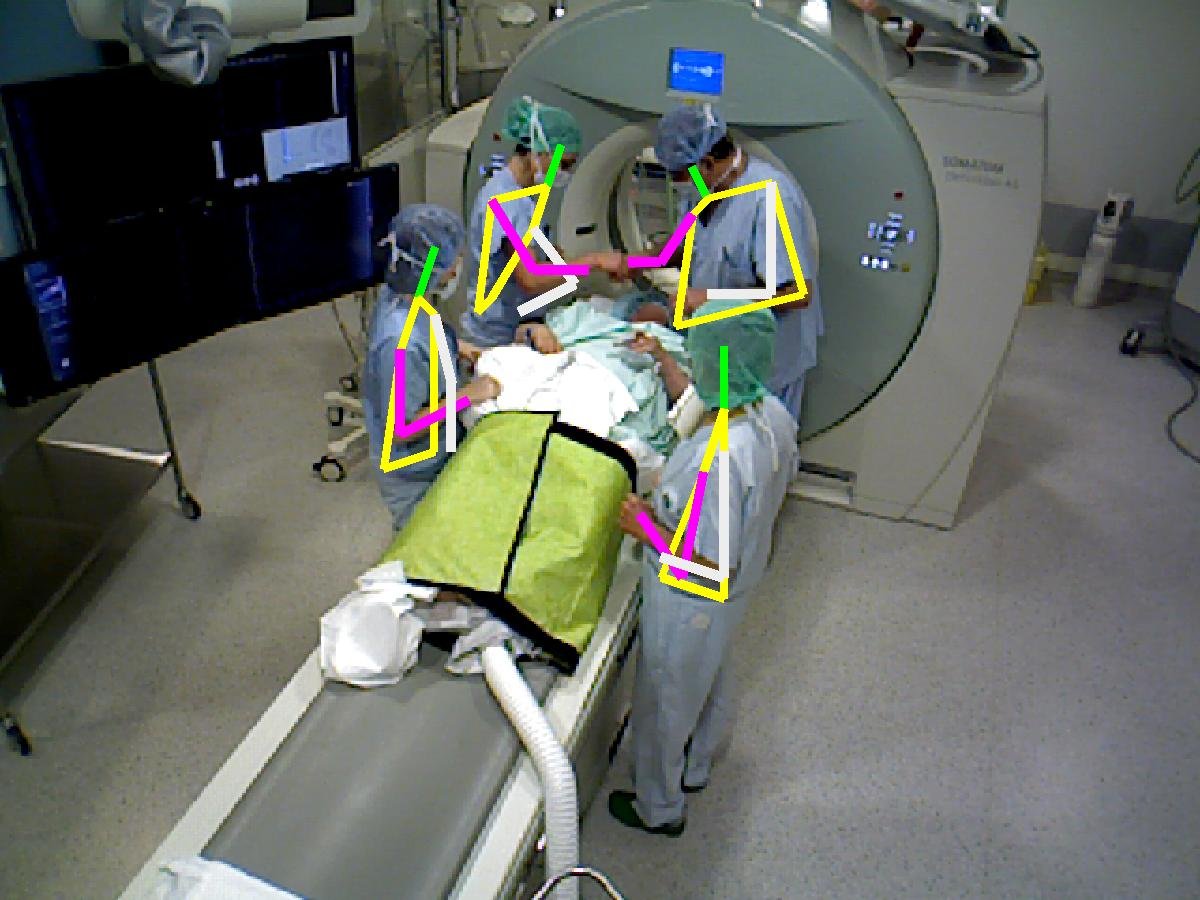}&
\includegraphics[width=0.24\linewidth]{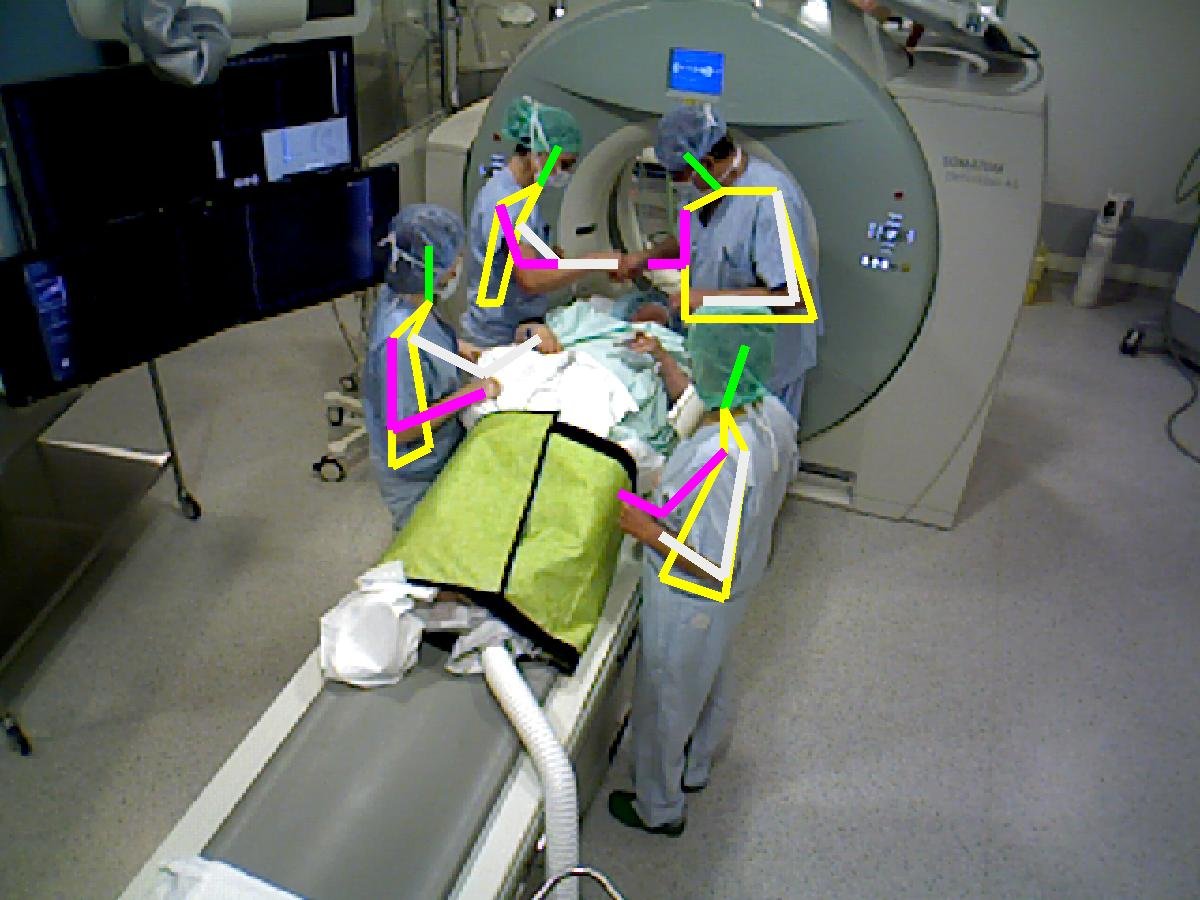} \\
{\small
$\psi_{2D}$ (I-HOG)} & {\small $\psi_{2D}$ (I-HOG+HDD)} & {\small $\psi_{3D}^4$ (I-HOG)} & {\small $\psi_{3D}^4$ (I-HOG+HDD)} 
\end{tabular}

\caption{Examples of pose estimation results for two different appearance models combined with two different pairwise constraints. White and magenta represent left and right arms, respectively. (Picture best seen in color)}
\label{fig:qualitRes}
\end{figure*}

Following common practice in the literature, we use the probability of correct keypoint (PCK) as evaluation criteria~\cite{yang_PAMI2013}. Given a tight bounding box for each person, PCK computes the probability of ground-truth joints to be localized with an error less than $\alpha.max(w,h)$, where $w$ and $h$ are the width and height of the bounding box, and $\alpha$ is $0.2$ as suggested by~\cite{yang_PAMI2013}. During training, we use ground-truth labels to cluster training data for constructing the mixtures. This step can be performed either in 2D or in 3D. We noticed in the experiments that clustering in 3D reduces the performance (by $\sim 5\%$ for $\psi_{3D}^{\{1-3\}}$). We believe that this is due to two reasons: {\it (1) noisy depth:} when the depth value for a ground-truth point is noisy, the 3D re-projection will be inaccurate. Therefore, noisy clusters are generated that lead to an inaccurate division of the part samples. (2) {\it insufficient number of samples for 3D clustering:} to avoid coarse clustering of the larger 3D space, more clusters are needed. However, increasing the number of clusters results in a smaller number of samples per mixture, leading in turn to weaker part detectors. Hereafter, we therefore report  the results when the part types are generated using 2D clustering. We perform our experiments on an Intel i7 machine, with six cores running at 3.20 GHz and 64 GB RAM. We implement our approach in MATLAB. In average, it takes five seconds and twelve seconds to test one image using our approach with a 2D deformation model and a 3D deformation model, respectively. However, if we do not use Algorithm~\ref{alg:buildingNG}, the computation with a 3D deformation model is 15 times slower. Even though the complexity remains quadratic in the size of the state space, these results show that the proposed inference approach speeds up the running time drastically. 

\begin{table}[t]
\centering
\setlength{\tabcolsep}{3pt} 
\begin{tabular}{|c c||c: c c c c|}
		\hline
			\multicolumn{2}{|c||}{Descriptor} & 2D & \multicolumn{4}{c|}{3D} \\
			Color &  Depth & $\psi_{2D}$ & $\psi_{3D}^1$& $\psi_{3D}^2$& $\psi_{3D}^3$& $\psi_{3D}^4$  \\
			\hline
	I-HOG & -& \ 63.3* & 64.8 & 38.0 & 56.0 & \textbf{69.3}\\
  - & D-HOG& 72.5 & 66.6 & 37.0 & 61.7 & \textbf{76.5} \\
  I-HOG & D-HOG& 75.3 & 73.6 & 58.4 & 66.6 & \textbf{79.5}\\
   \hline
  - & HONV	&  65.6 & 67.3 & 39.5 & 54.4 &\textbf{71.0}\\
  I-HOG & HONV& 75.4 & 72.9 & 55.0 & 68.8 & \textbf{80.1}\\
  \hline
  - & HDD  &  74.7 & 73.0 & 46.7 & 67.7 & \textbf{79.1} \\
  I-HOG& HDD& 76.6 & 76.6 & 70.3 & 72.3 &\textbf{81.5}\\
		\hline
	\end{tabular}
	\caption{PCK results. Comparison of 5 deformation models in combination with seven different appearance models. Each row shows the evaluation results for an appearance model in combination with the 2D pairwise constraint $\psi_{2D}$ or one of the proposed 3D pairwise constraints $\psi_{3D}^{1-4}$ as deformation model. Note(*): $\psi_{2D}$ with I-HOG is FMP~\cite{yang_PAMI2013}. }
	\label{tab:comparisonAvg}
\end{table}

\begin{figure*}[t]
\setlength{\tabcolsep}{1pt} 
\renewcommand{\arraystretch}{0.5} 

\begin{tabular}{ccccc}
\includegraphics[width=0.19\linewidth]{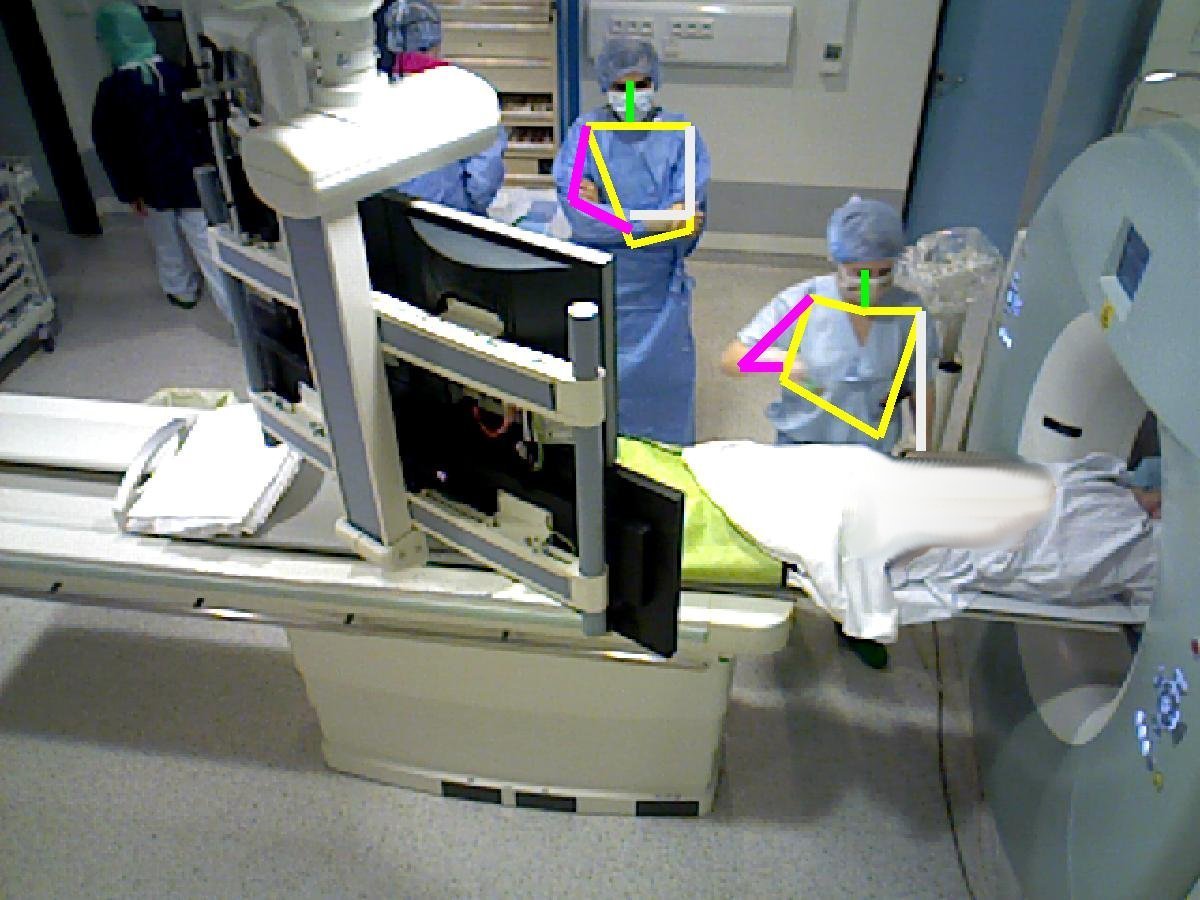} &
\includegraphics[width=0.19\linewidth]{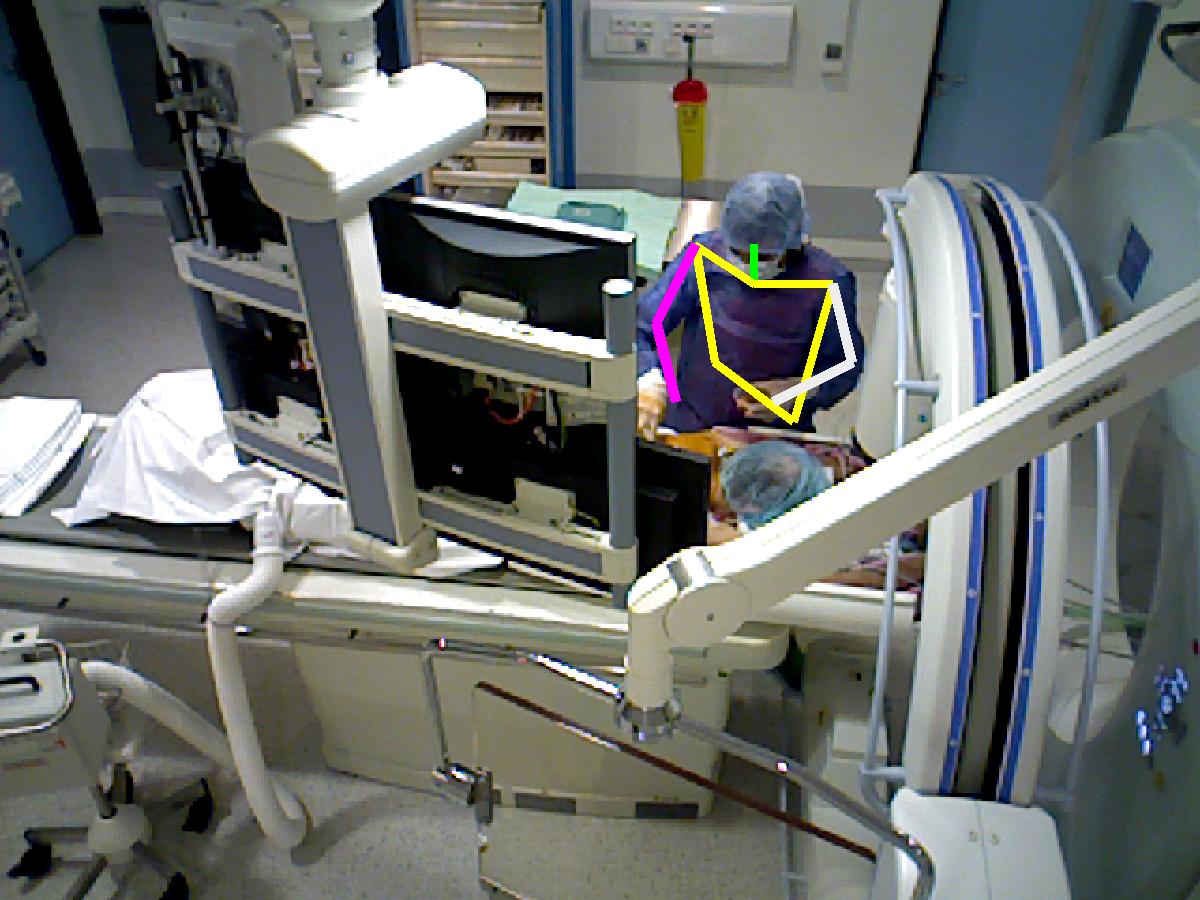} &
\includegraphics[width=0.19\linewidth]{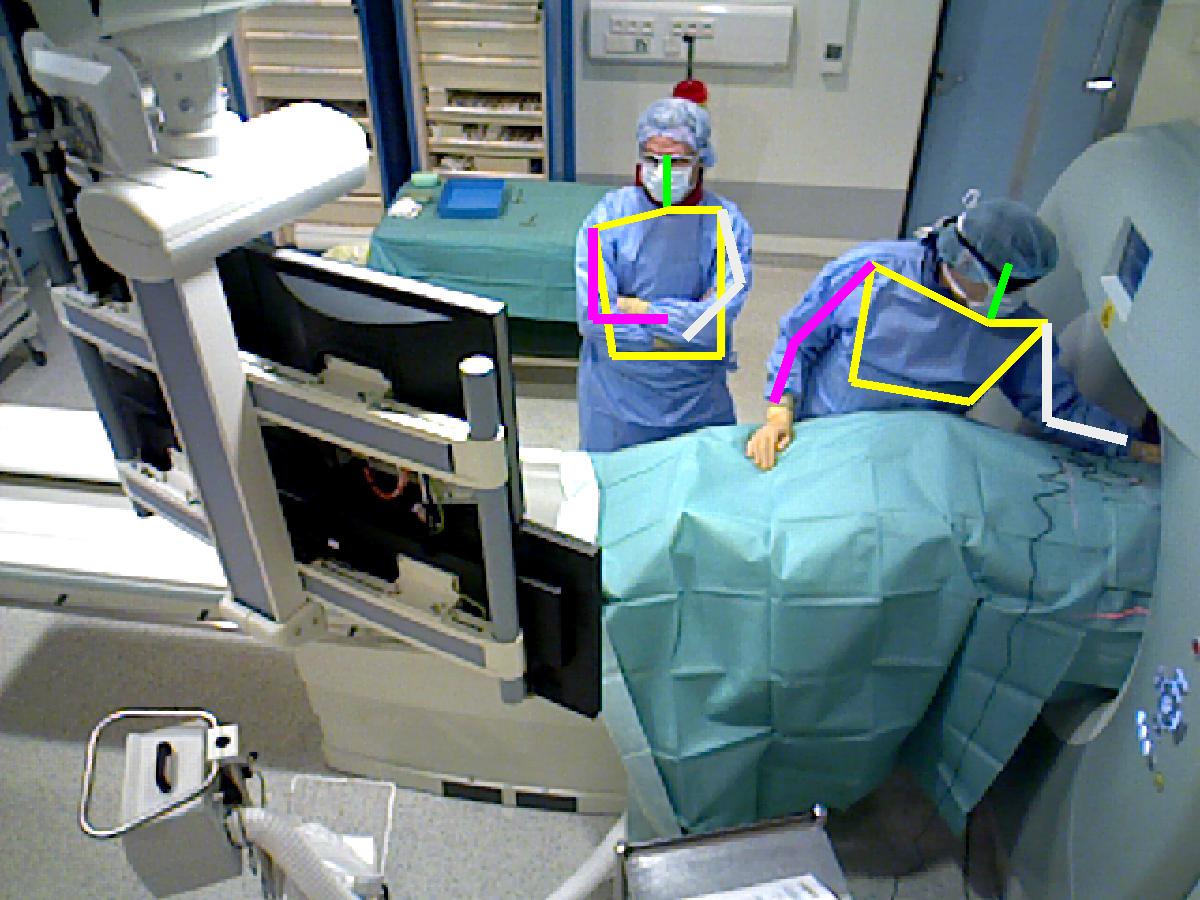} & \includegraphics[width=0.19\linewidth]{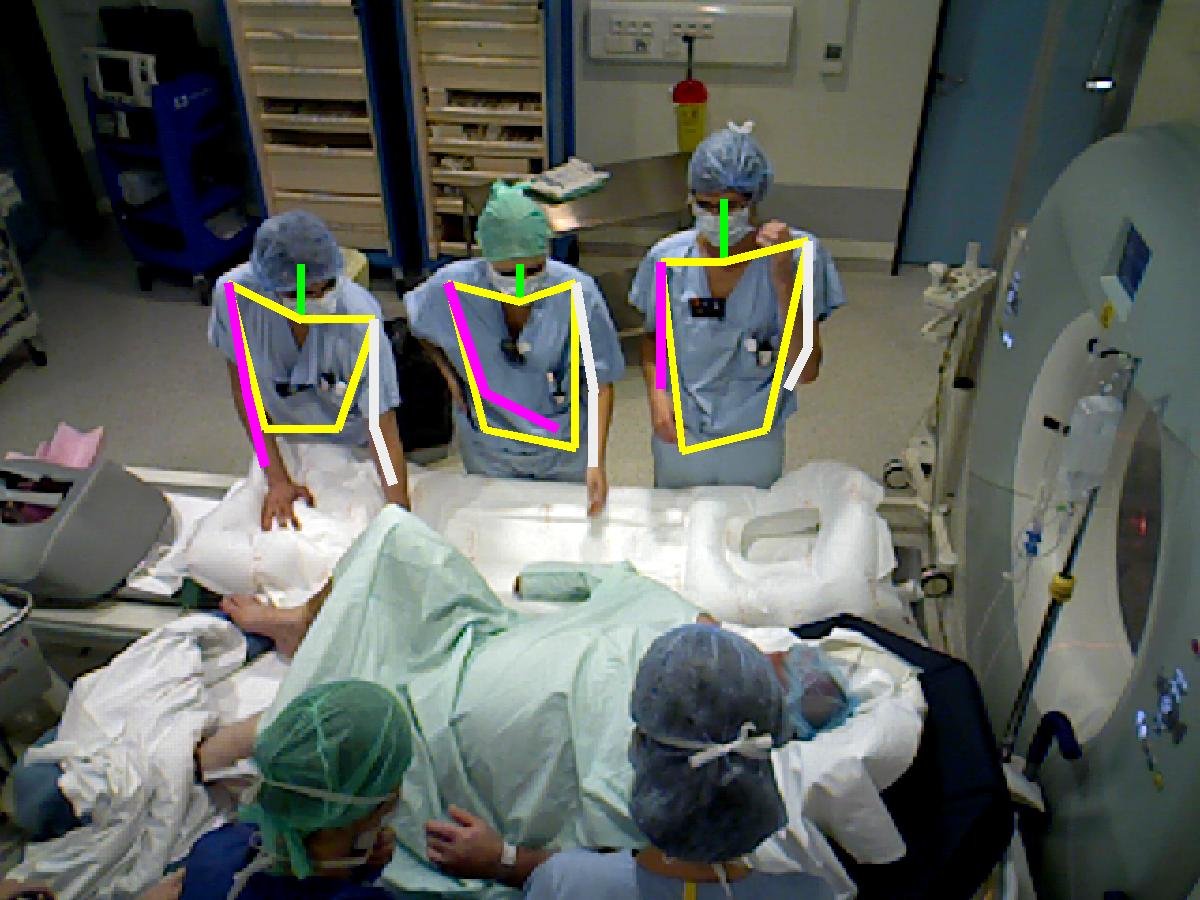} &
\includegraphics[width=0.19\linewidth]{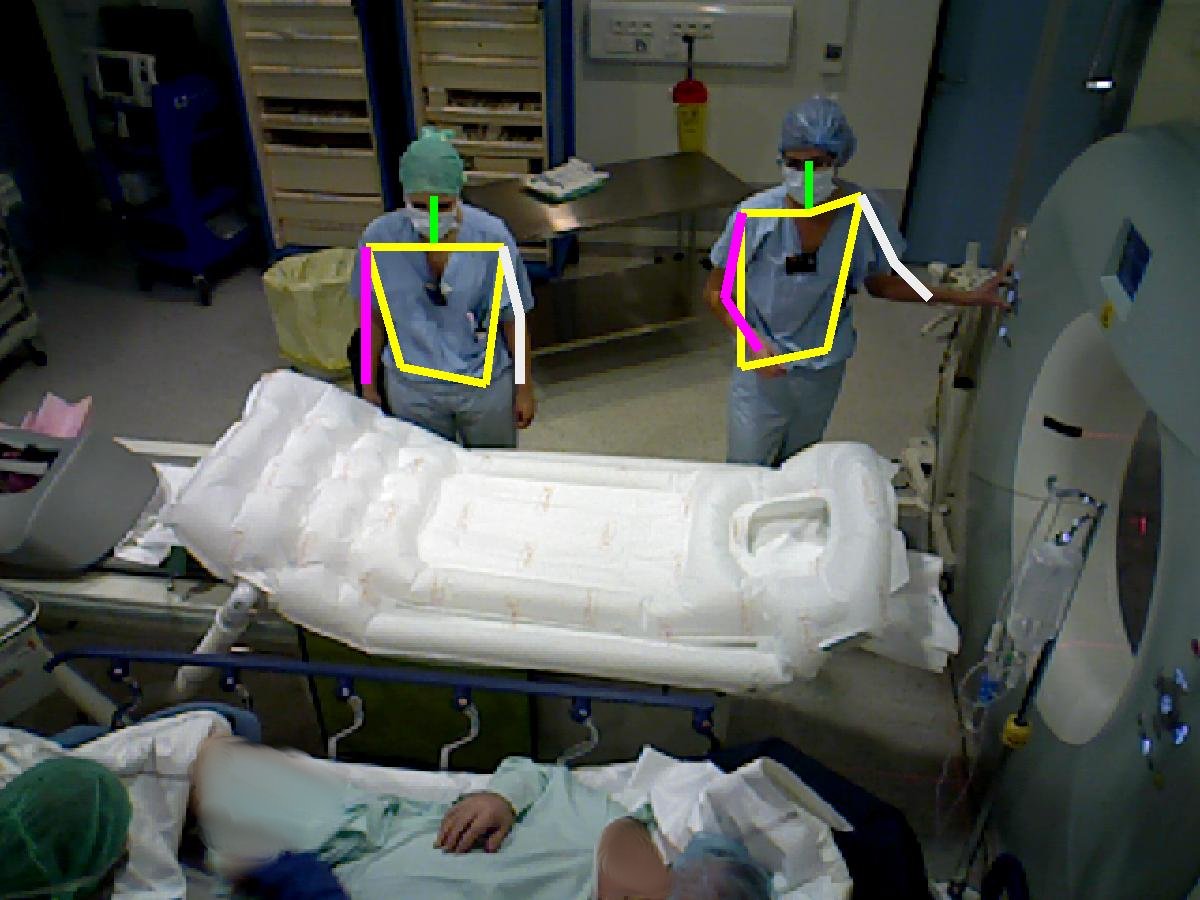} \\
\includegraphics[width=0.19\linewidth]{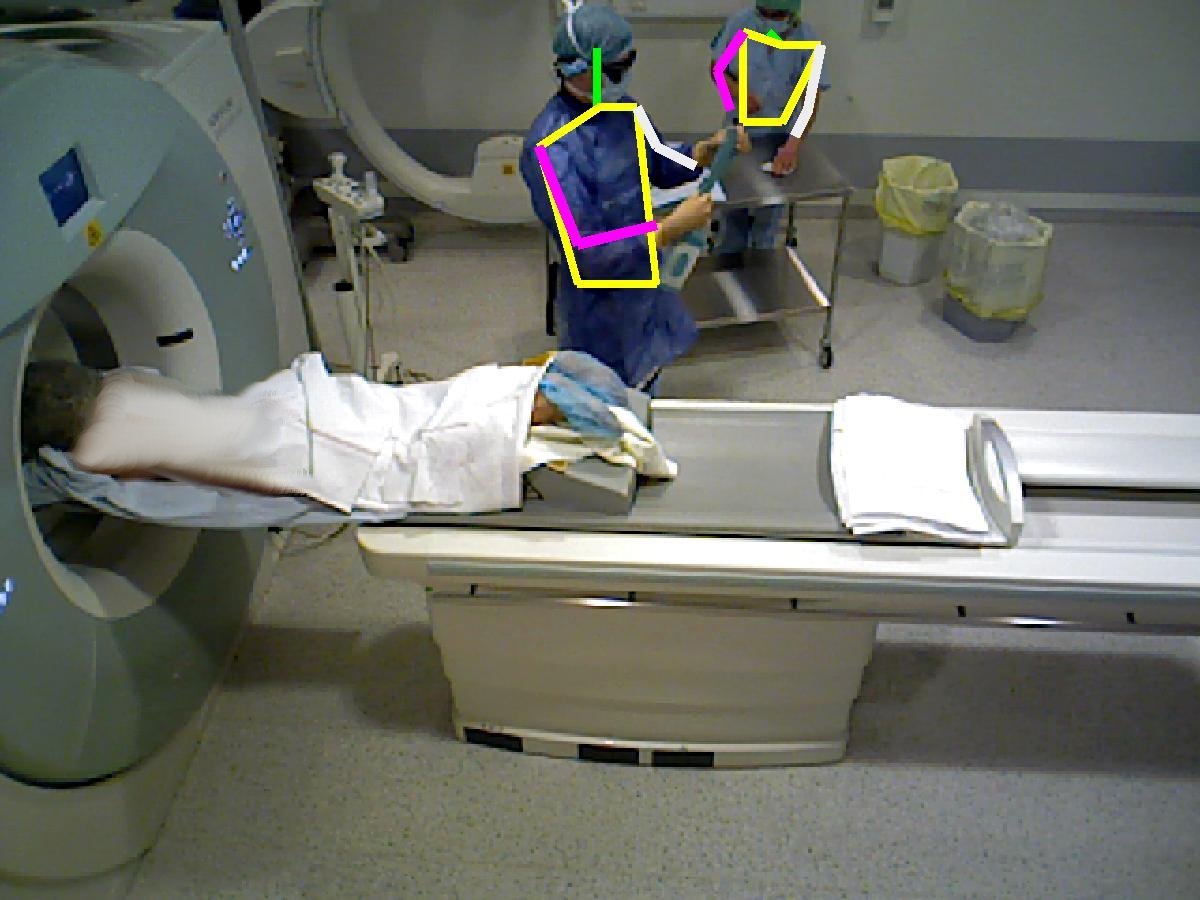} &
\includegraphics[width=0.19\linewidth]{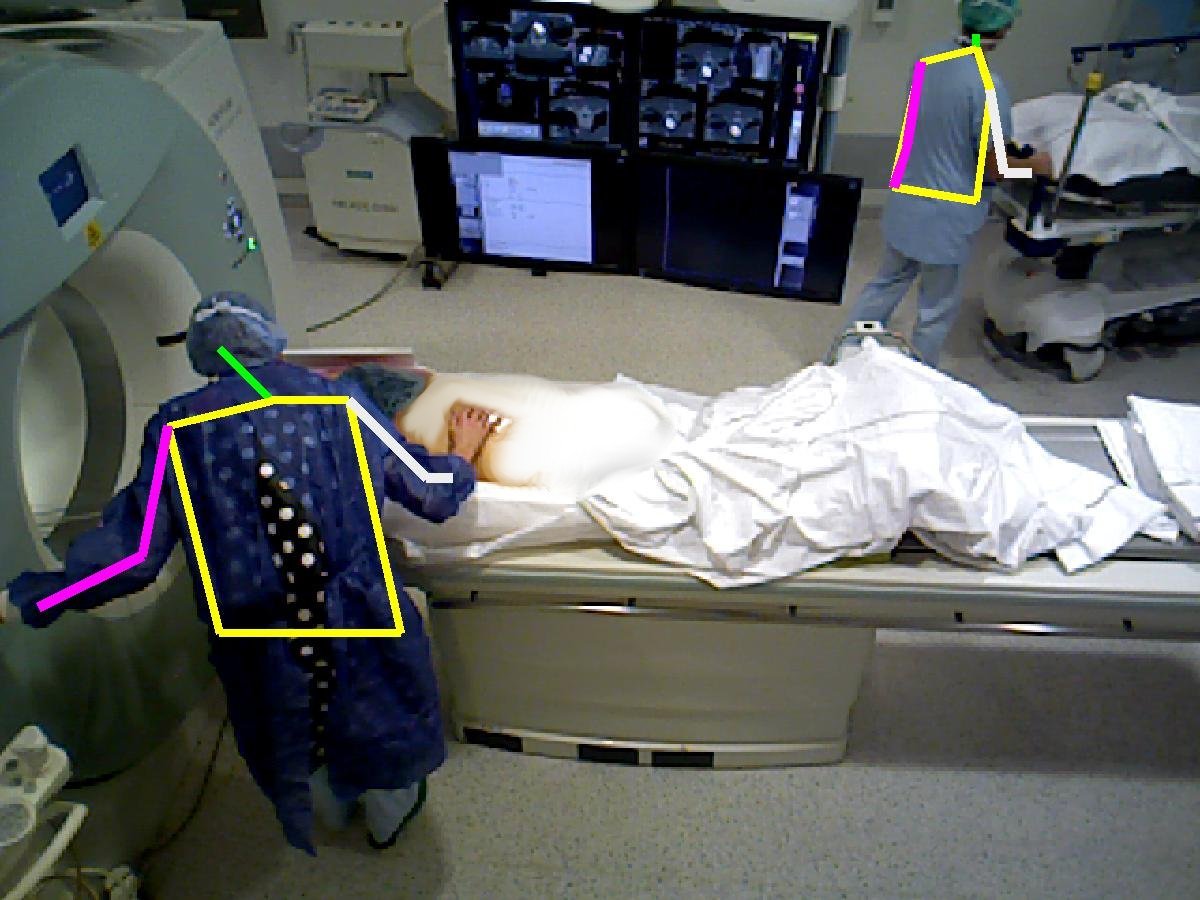} &
\includegraphics[width=0.19\linewidth]{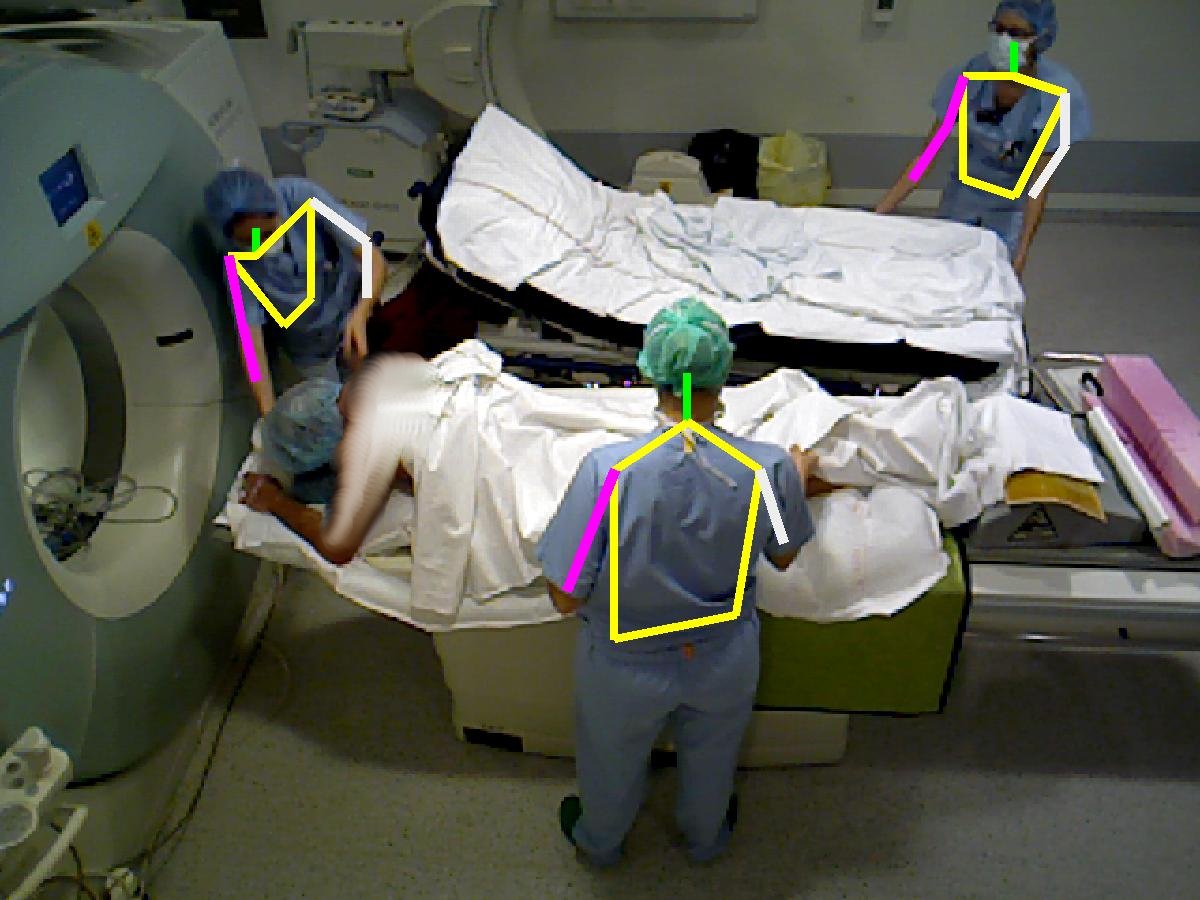} & \includegraphics[width=0.19\linewidth]{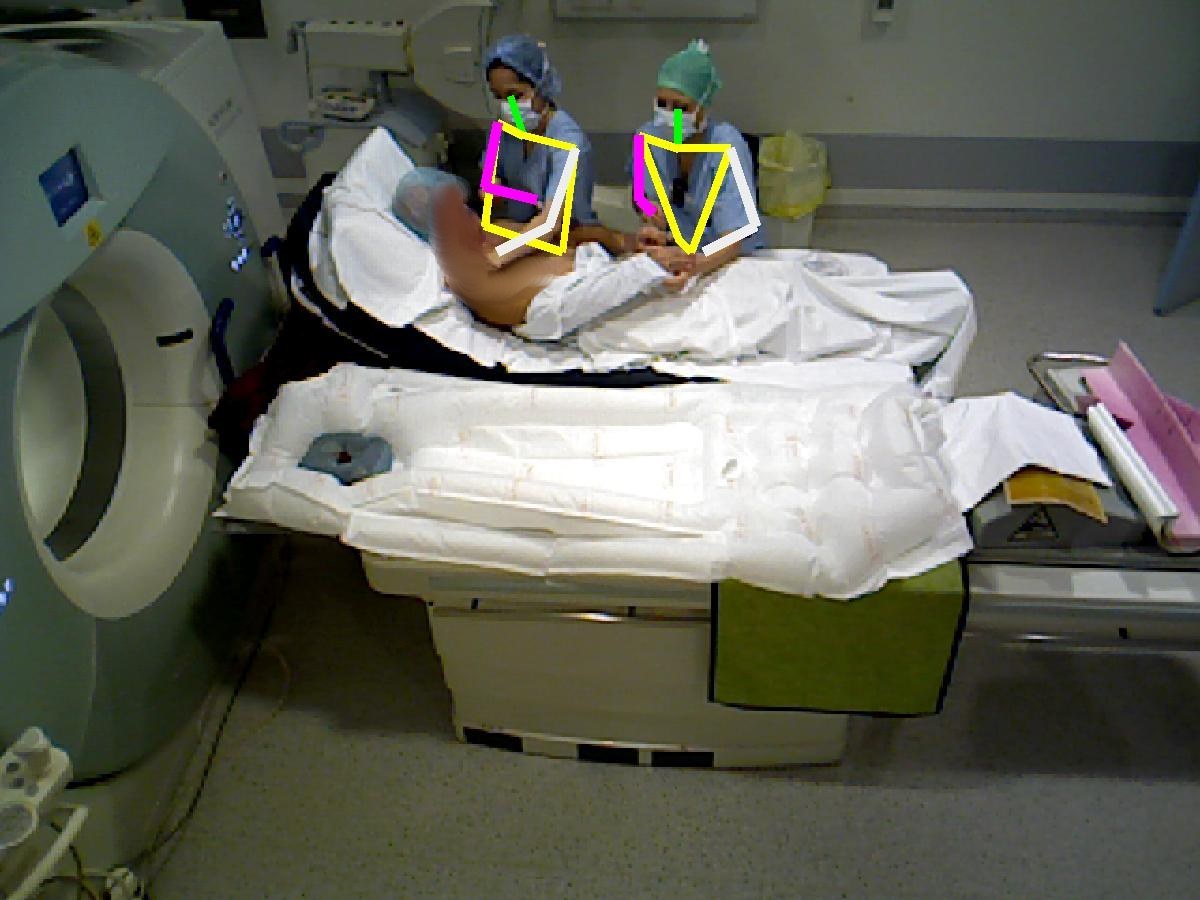} &
\includegraphics[width=0.19\linewidth]{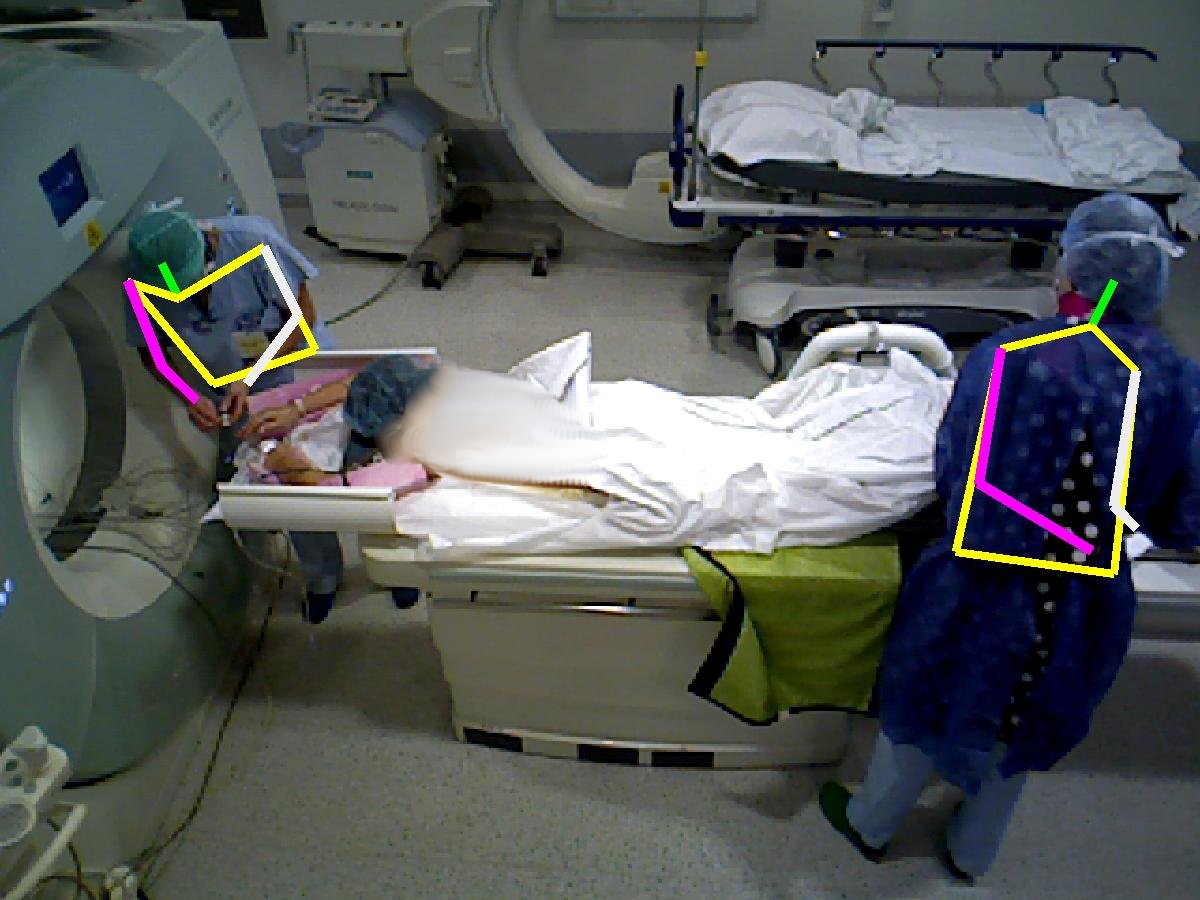} \\
\includegraphics[width=0.19\linewidth]{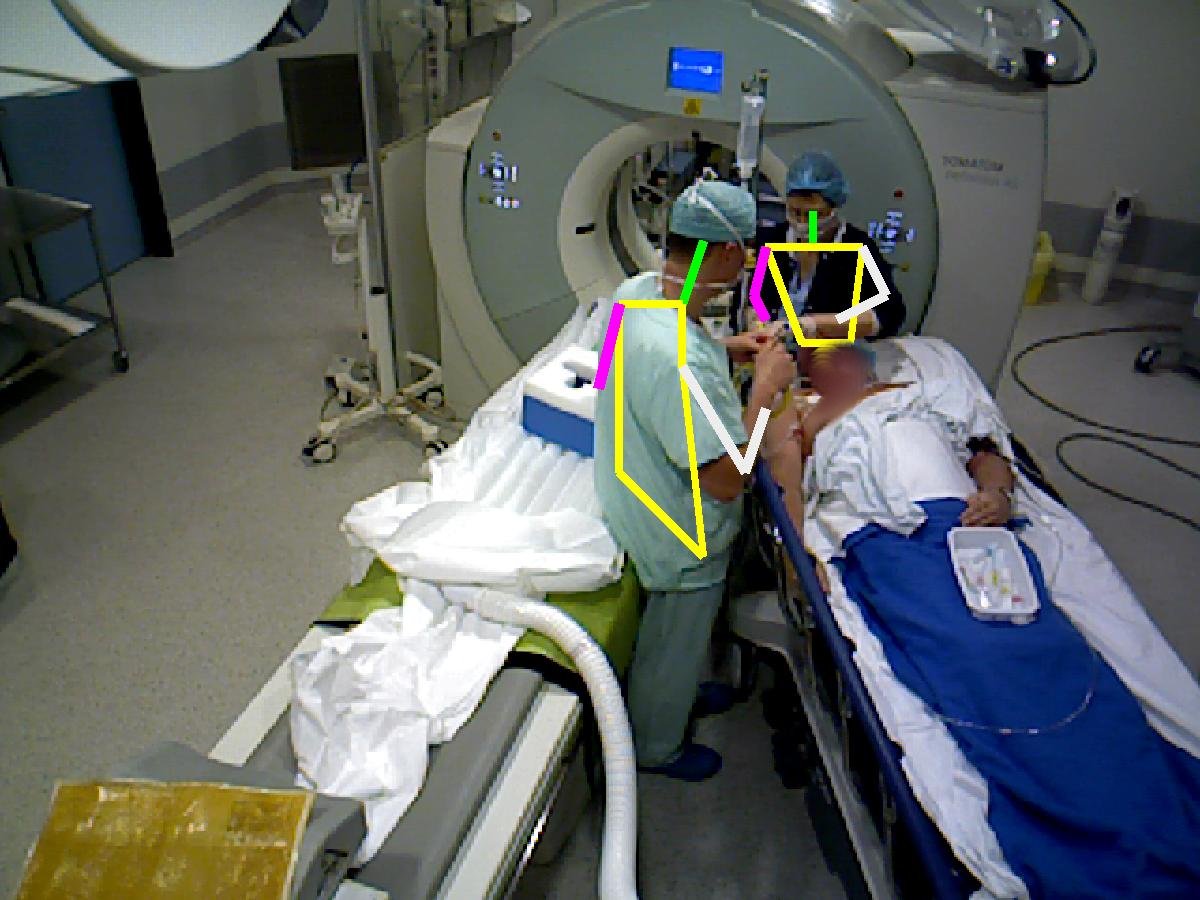} &
\includegraphics[width=0.19\linewidth]{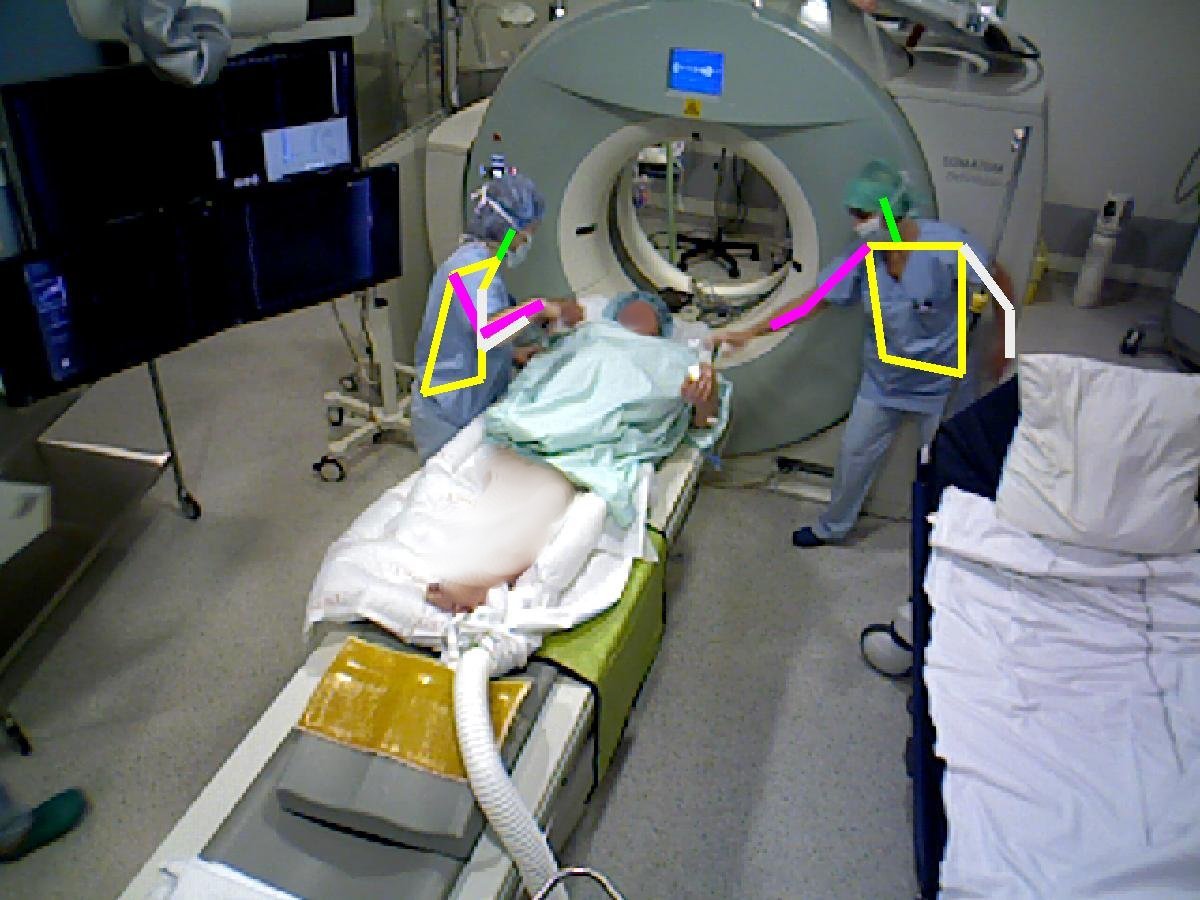} &
\includegraphics[width=0.19\linewidth]{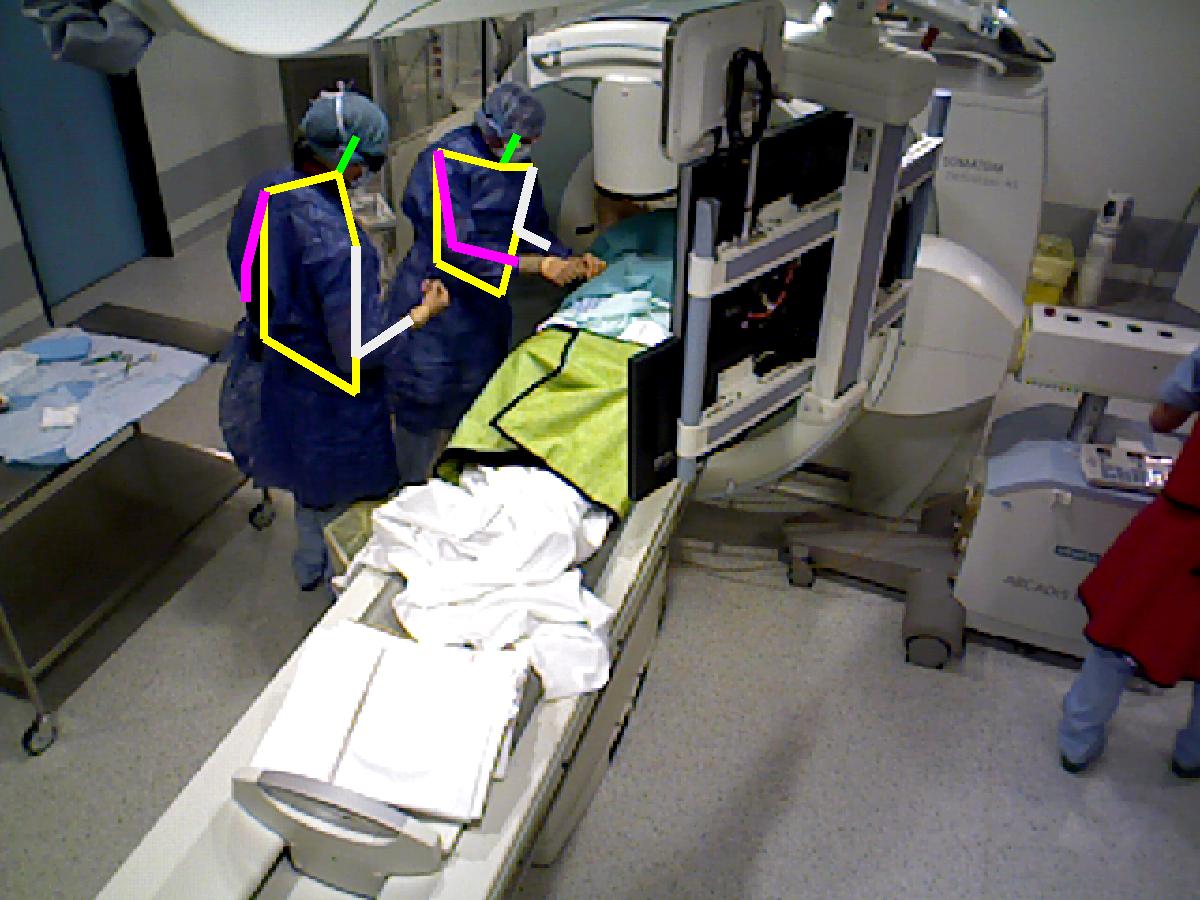} & \includegraphics[width=0.19\linewidth]{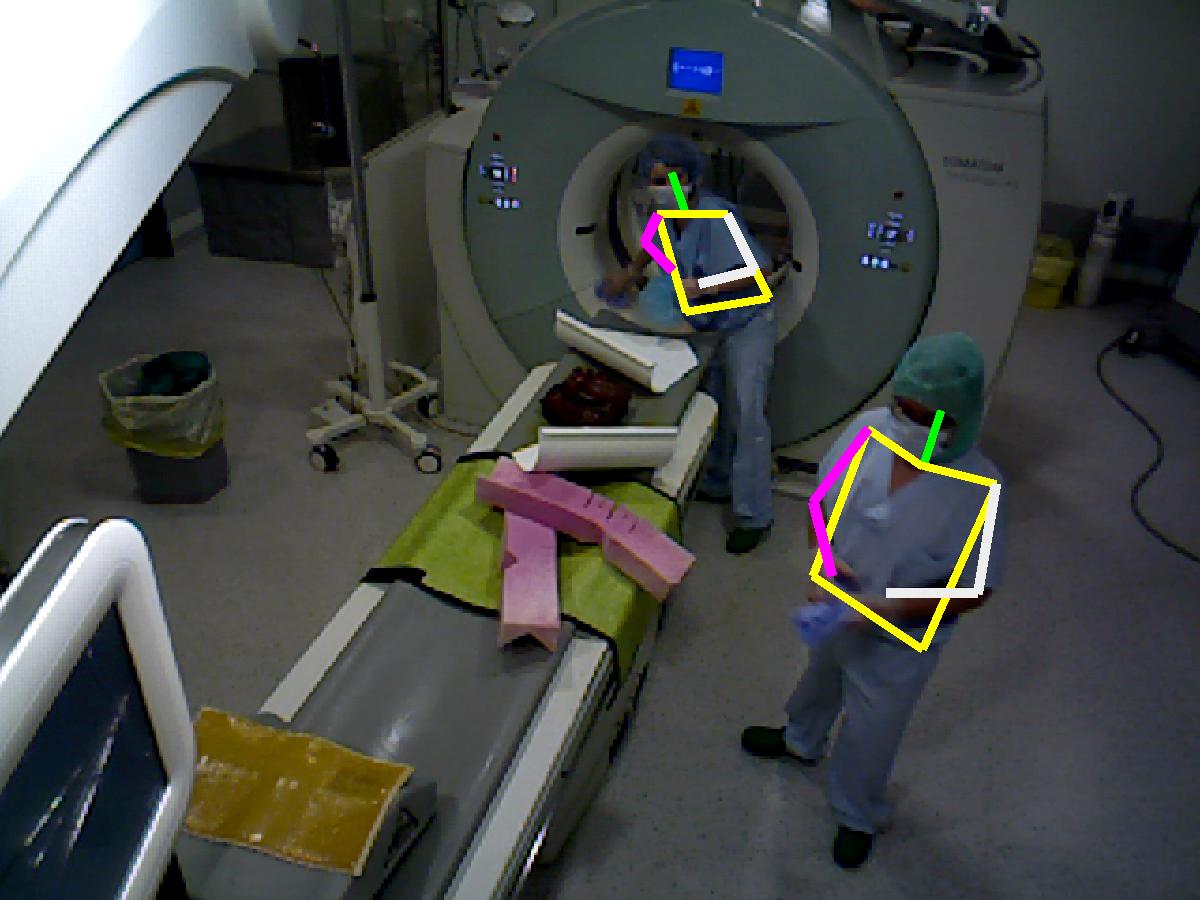} &
\includegraphics[width=0.19\linewidth]{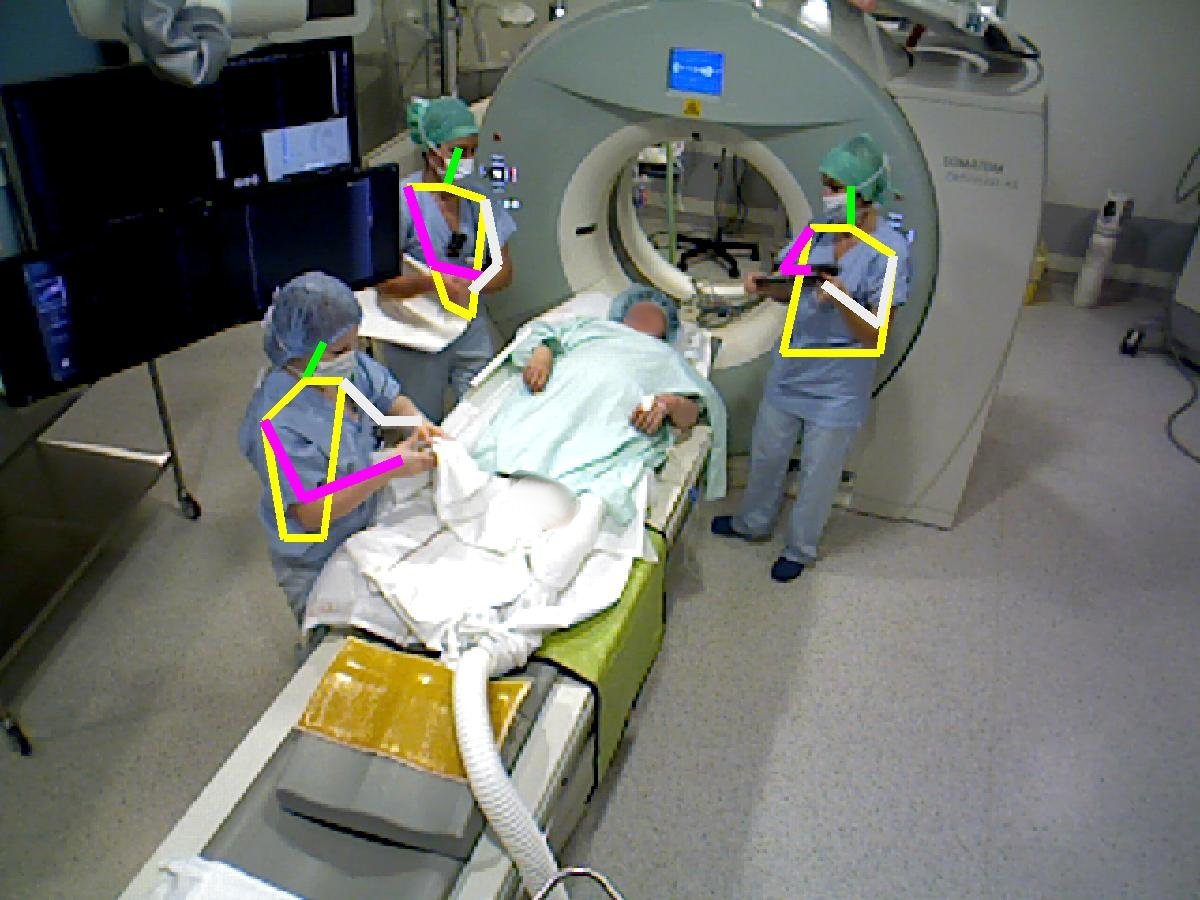} \\
\end{tabular}
\caption{Examples of pose estimation results obtained with the proposed 3D pictorial structures approach using $\psi_{3D}^4$ with I-HOG+HDD. (Picture best seen in color)}
\label{fig:qualitResPsi4}
\end{figure*}

Table~\ref{tab:comparisonAvg} presents the average performance results of the seven-fold cross validation. The results of different appearance models in comparison with the FMP approach~\cite{yang_PAMI2013} on the same experimental setup are presented in column \lq 2D\rq. FMP uses HOG on color images and 2D pairwise constraints. The results show that the representation based on D-HOG significantly improves the performance over FMP. Considering that both I-HOG and D-HOG are using the same descriptor, it indicates that the depth gradient is more reliable than the color-based one in environments with high color similarities and illumination changes. In general, the depth-based appearance models always outperform the color-based one and the best performance is obtained by HDD. The HDD based appearance model improves the performance over the baseline by $\sim11\%$. This highlights the benefit of the proposed coarse and depth invariant representation in describing surface level changes. One can also notice that combining a depth-based descriptor with I-HOG always boosts the performance by building stronger appearance models that make use of complementary information coming from both color and depth images. For the sake of comparison, we have also built a depth appearance model that combines all depth descriptors, namely D-HOG, HONV and HDD. However, this model does not yield any significant improvement. 

The results for the proposed 3D pairwise constraints in combination with different appearance models are also presented in Table~\ref{tab:comparisonAvg}. In general, using $\psi^{\{1-3\}}_{3D}$ does not improve the performance. We believe that this is due to the noisy depth coming from the affordable RGB-D camera. The noisy depth leads to inaccurate ground-truth projection into 3D, which in turn results in  incorrect part length estimations and relative displacements $\left(\overline{dx}, \overline{dy}, \overline{dz}\right)$. Moreover, since all parameters are learned in a unified framework, it will not only affect the deformation model, but also the part detectors. These inaccurate part lengths have more impact on $\psi^2_{3D}$ that uses both the absolute 3D Euclidean distance and the magnitudes of relative 3D displacements between parts along the axes.

The best performance is always obtained with $\psi^4_{3D}$, which significantly improves the performance over the model with 2D pairwise constraints using the same descriptor. The highest performance boost of 6\% is obtained for the appearance model that only relies on color  and does not use any depth information (FMP). Due to the high color similarity in the images, the part detector provides noisy detections. The 2D deformation model is not able to resolve the uncertainty caused by these weak detections, contrary to the proposed 3D deformation model. These results show that by using more reliable pairwise dependencies, the PS model can better resolve the uncertainty of the part detector. These improvements by $\psi^4_{3D}$ also demonstrate that this pairwise term provides an elegant way to benefit from the 3D distances to learn a more reliable deformation model and to use the 2D positions to be more robust to the noise present in the re-projected 3D positions.

Figure~\ref{fig:qualitRes} shows the estimated poses using $\psi_{2D}$ with I-HOG (\ie FMP), $\psi_{2D}$ with I-HOG+HDD, $\psi^4_{3D}$ with I-HOG and $\psi^4_{3D}$ with I-HOG+HDD. The results show that 2D PS  is confused by false detections on the background and also mixes up detections between persons. The last row shows cases where the 3D PS approach does not localize the arms correctly, although the heads and shoulders are correctly estimated. It is either due to weak part detection responses, occlusions or side view poses. Figure~\ref{fig:qualitResPsi4} shows more qualitative results obtained by the proposed 3D pictorial structures that uses I-HOG+HDD and $\psi_{3D}^4$. In summary, the use of 3D information always boosts the performance, when used in the appearance model alone, in the deformation model alone or in both. The best results are obtained when 3D information is used in both models.


\begin{table}[t]
\centering
\setlength{\tabcolsep}{7pt} 
\begin{tabular}{|l|c c :c c |}
		\hline
			Body parts  & \multicolumn{2}{c:}{I-HOG} & \multicolumn{2}{c|}{I-HOG+HDD}\\
			 & $\psi_{2D}$ & $\psi_{3D}^4$ &$\psi_{2D}$ & $\psi_{3D}^4$\\
			\hline
			Head 	 & 84.1 & \textbf{92.3} & 92.8 & \textbf{96.4}\\
			Shoulder & 72.7 & \textbf{80.5} & 84.1 & \textbf{87.7}\\
			Elbow 	 & 57.0 & \textbf{59.7} & 71.1 & \textbf{76.6}\\
			Wrist 	 & 56.5 & \textbf{64.4} & 71.6 & \textbf{76.8}\\
			Hip 		 & 45.9 & \textbf{52.6} & 63.6 & \textbf{69.9}\\
			\hline
			Average	 & 63.3 & \textbf{69.3} & 76.6 & \textbf{81.5}\\
		\hline
	\end{tabular}
	\caption{PCK evaluation results per body part. Part detection for three variants of our approach compared with baseline FMP (I-HOG+$\psi_{2D}$)~\cite{yang_PAMI2013}.}
	\label{tab:pckResultsPerParts}
\end{table}

The detailed performance results per body part are presented in Table~\ref{tab:pckResultsPerParts}. We present the results for  $\psi_{2D}$ and $\psi^4_{3D}$ as well as for the I-HOG and I-HOG+HDD appearance models. 
 It is important to notice that 2D PS is using exact inference and that these results are therefore the best possible using these appearance models. The proposed 3D deformation model improves the results consistently and significantly.
 We therefore show that a reliable 3D deformation model permits to efficiently deploy PS on RGB-D data, while the experiments suggest that 2D-based deformation models are limited by their unreliable pixel-based distance metric.

\subsection{Clinician detection}

We use our clinician pose estimation approach to perform human detection in operating rooms as well. Detection windows are obtained by fitting tight bounding boxes around all estimated poses. We compare our approach with deformable part models (DPM)~\cite{felzenszwalb_pami2010}, which achieve competitive results on challenging datasets for human and object detection. The DPM approach uses multiple mixtures of multiscale deformable part models to perform object detection using ensembles of parts. The approach uses bounding box annotations to learn a set of root filters for each object and then uses the filters to divide the object into several parts. Finally, after defining the parts, an energy function similar to FMP is used for detection. The energy function consists of an appearance model, a deformation model and a mixture bias. To build a stronger baseline, we have also extended the appearance model in DPM to use both color and depth images. However, since bounding boxes are used to specify parts that include both  foreground and background, it is not straightforward to extend the deformation model to 3D. 

To evaluate clinician detection, we use the average precision (AP) score that is commonly used for object and human detection in the literature as well as in~\cite{felzenszwalb_pami2010}. A detection box is considered as true positive if it has more than 50\% overlap with a ground-truth bounding box. Multiple detections are penalized, \ie if more than one detection for a ground-truth occur, one detection will be accepted as a true positive and the others will be considered as false positives. This criteria is used to compute a precision-recall curve and AP is the area under the curve.     

\begin{table}[]
\setlength{\tabcolsep}{3pt}
\centering
\begin{tabular}{|c|c|c|c|c|c|c|}
\hline
          & \multicolumn{2}{c|}{DPM} & \multicolumn{2}{c|}{PS($\psi_{2D}$)}    & \multicolumn{2}{c|}{PS($\psi_{3D}^{4}$)} \\ \cline{2-7} 
          & N            & N+D           & N             & N+D             & N           & N+D          \\ \hline
I-HOG     & 75.5         & 70.0        & 68.8         & 64.0        & 77.3          & 72.0          \\ \hline
I-HOG+HDD & 80.3        & 75.1       & 86.8         & 79.1        & \textbf{89.7} & \textbf{80.8} \\ \hline
\end{tabular}
\caption{Person detection results using AP score. Two variants of our approach are compared with DPM on the same appearance models. N indicates a set of annotated staff who have at least half of their upper body visible in the view. N+D contains all annotated staff appearing in the view.}
\label{tab:perDet}
\end{table}

\begin{figure}[t]
\includegraphics[width=.95\columnwidth]{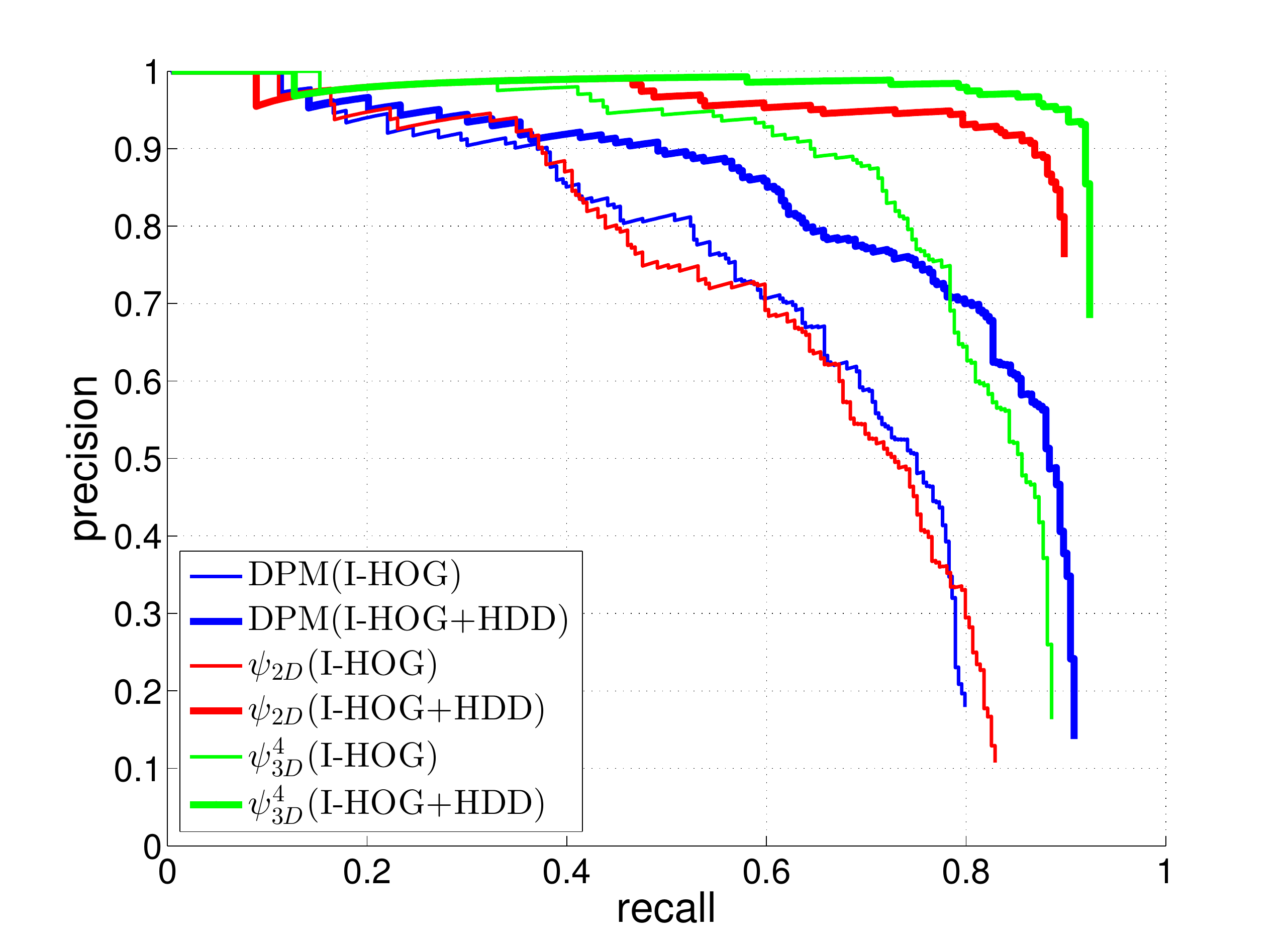}
\caption{Precision-recall curves computed for the detection of normal staff in the first fold of the cross validation. Results for DPM, $\psi_{2D}$ and $\psi_{3D}^4$ in combination with the I-HOG and I-HOG+HDD representations.}
\label{fig:preRec}
\end{figure}

Table~\ref{tab:perDet} presents clinician detection results. In this table, we present the results for two appearance models: I-HOG that is used in both FMP~\cite{yang_PAMI2013} and DPM~\cite{felzenszwalb_pami2010}, and I-HOG+HDD that was the best appearance model according to our experiments for clinician pose estimation. Following the same reasoning, we also use the two pairwise constraints  $\psi_{2D}$ and $\psi_{3D}^4$. For clinician detection, we consider two cases: {\it (1) Normal staff:} we compute the true/false positives and negatives only for staff that are labeled as normal, indicated by N in the table. The first detection for a difficult staff is not considered as false positive. If a staff with difficult flag does not have a detection, it is not considered as a false negative. {\it (2) Normal and Difficult staff:} Missing detection of any staff is considered as false negative, indicated by N+D in the table. 

The representation based on I-HOG+HDD boosts the performance for DPM by $\sim5\%$. These consistent improvements indicate that the jointly learned appearance model is highly beneficial in visually challenging environments for both task of human pose estimation and detection. Figure~\ref{fig:preRec} summarizes clinician detection results on normal staff using precision-recall curves computed on the first fold. The increase in the precision and recall of the approaches based on the proposed I-HOG+HDD representation indicates the benefits of this representation in building more reliable and discriminative models. Furthermore, this representation enables all models, namely DPM, $\psi_{2D}$ and $\psi_{3D}^4$ to obtain a similar maximum recall value. The high precision for $\psi_{3D}^4$ also highlights the advantages of the 3D deformation model in pruning false positives.

The original DPM (I-HOG) outperforms standard FMP on clinician detection for two reasons. First, since the number of bounding box annotations is much higher than the number of pose annotations, DPM has access to a larger training set to learn the model. Second, DPM clusters the training data based on box sizes to learn different components. Since the box sizes are normally changing according to the distance of the person to the camera, this enables DPM to learn different components for people at different distances.
However, one can notice that by using the robust and discriminative I-HOG+HDD representation, which can encode 3D information, our clinician detection approach consistently outperforms DPM. The best clinician detection result is obtained using our clinician detection approach with  deformation model $\psi_{3D}^4$. These results further indicate the benefits of 3D pairwise constraints for human detection in cluttered and crowded scenes.

\section{Conclusions}
This paper proposes an approach based on pictorial structures for human pose estimation and detection in operating rooms. By designing appearance models based on both color and depth images as well as deformation models based on 3D pairwise constraints, we extend PS to 3D on RGB-D data. We also present a new feature for depth images, the histogram of depth differences,  which encodes surface level changes in a coarse, multi-scale and depth invariant representation. Finally, we quantitatively evaluate the approach on a novel and challenging dataset generated from several days of recordings during live surgeries. Different combinations of the proposed appearance and deformation models  are compared to state-of-the-art methods for human pose estimation~\cite{yang_PAMI2013} and human detection~\cite{felzenszwalb_pami2010}. Our results highlight the strength of the proposed appearance model, where the best performance is always obtained by I-HOG+HDD. Furthermore, experimental results demonstrate how the 3D pairwise constraints significantly improve the results for both clinician detection and pose estimation in a cluttered and busy environment like the OR. Key to this improvement is the use of 3D information to (1) construct 3D nodes; (2) reduce the number of edges in the connectivity map of the state space; and (3) propagate information in the state space by considering 3D distances between the nodes, while retaining an exact solution. 
To the best of our knowledge, this is the first time that an approach for articulated clinician detection is proposed and evaluated on a large dataset recorded during real surgeries. 
In future work, we plan to investigate how to extend this approach to multiple views in order to improve the detections in cluttered scenes.

\section*{Acknowledgements}

This work was supported by French state funds managed by the ANR within
the Investissements d'Avenir program under references ANR-11-LABX-0004
(Labex CAMI), ANR-10-IDEX-0002-02 (IdEx Unistra) and ANR-10-IAHU-02
(IHU Strasbourg).

\bibliographystyle{IEEEtran}
\bibliography{IEEEabrv,3DPSforClinicianDetection}


\end{document}